\pdfoutput=1
\documentclass[letterpaper]{article} 
\usepackage{aaai24}  
\usepackage{times}  
\usepackage{helvet}  
\usepackage{courier}  
\usepackage[hyphens]{url}  
\usepackage{graphicx} 
\usepackage{natbib}  
\usepackage{caption} 
\usepackage{amssymb}
\usepackage{multirow}
\usepackage{makecell}
\usepackage{booktabs}
\usepackage{siunitx}
\usepackage{blindtext}
\usepackage{multicol}
\usepackage[pagebackref,breaklinks,colorlinks]{hyperref}

\usepackage{float}
\usepackage{subfig}
\usepackage{multirow}
\usepackage{booktabs}
\usepackage{color, colortbl}
\usepackage{makecell}
	
\definecolor{Gray}{gray}{0.9}

\usepackage{newfloat}
\usepackage{listings}
\DeclareCaptionStyle{ruled}{labelfont=normalfont,labelsep=colon,strut=off} 
\floatstyle{ruled}
\newfloat{listing}{tb}{lst}{}
\floatname{listing}{Listing}
\title{MLLM-DataEngine: An Iterative Refinement Approach for MLLM} 

\author{
   Zhiyuan Zhao$^{*}$, 
   Linke Ouyang$^{*}$, 
   Bin Wang$^{*}$,  \\
   Siyuan Huang, 
   Pan Zhang, 
   Xiaoyi Dong, 
   Jiaqi Wang, 
   Conghui He$^{\dag}$
}
\affiliations{
    Shanghai AI Laboratory\\
    \{zhaozhiyuan,ouyanglinke,wangbin,huangsiyuan1,zhangpan,dongxiaoyi,wangjiaqi,heconghui\}@pjlab.org.cn 
}

\usepackage{bibentry}

\begin{document}

\maketitle

\begin{abstract}
Despite the great advance of Multimodal Large Language Models (MLLMs) in both instruction dataset building and benchmarking, the independence of training and evaluation makes current MLLMs hard to further improve their capability under the guidance of evaluation results with a relatively low human cost. In this paper, we propose MLLM-DataEngine, a novel closed-loop system that bridges data generation, model training, and evaluation. Within each loop iteration, the MLLM-DataEngine first analyze the weakness of the model based on the evaluation results, then generate a proper incremental dataset for the next training iteration and enhance the model capability iteratively. Compared with previous data collection methods which are separate from the benchmarking, the data generated by MLLM-DataEngine shows better targeting, quality, and correctness. For targeting, we propose an Adaptive Bad-case Sampling module, which adjusts the ratio of different types of data within each incremental dataset based on the benchmarking results. For quality, we resort to GPT-4 to generate high-quality data with each given data type. For correctness, prompt design is critical for the data generation results. Rather than previous hand-crafted prompt, we propose an Interactive Prompt Optimization strategy, which optimizes the prompt with the multi-round interaction between human and GPT, and improve the correctness of generated data greatly. Through extensive experiments, we find our MLLM-DataEngine could boost the MLLM capability in a targeted and automatic manner, with only a few human participation. We hope it could be a general solution for the following MLLMs building. The MLLM-DataEngine has been open-sourced and is now available at \url{https://github.com/opendatalab/MLLM-DataEngine}. 


\end{abstract}

\begin{figure*}[h]
    \centering
    \subfloat[ChatCaptioner]{\includegraphics[scale=0.35]{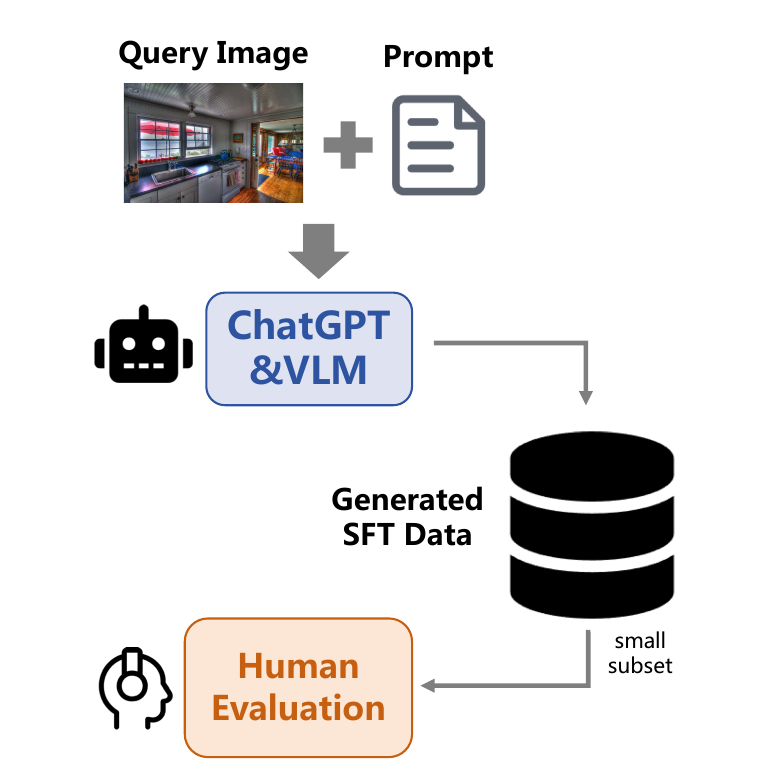}}
    \subfloat[IdealGPT]{\includegraphics[scale=0.35]{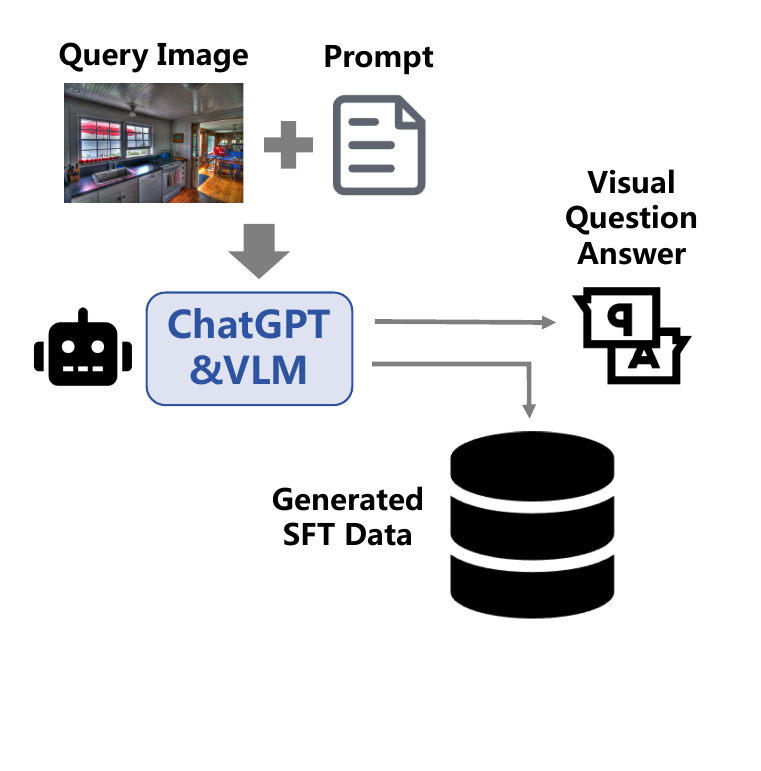}}
    \subfloat[LLAVA]{\includegraphics[scale=0.35]{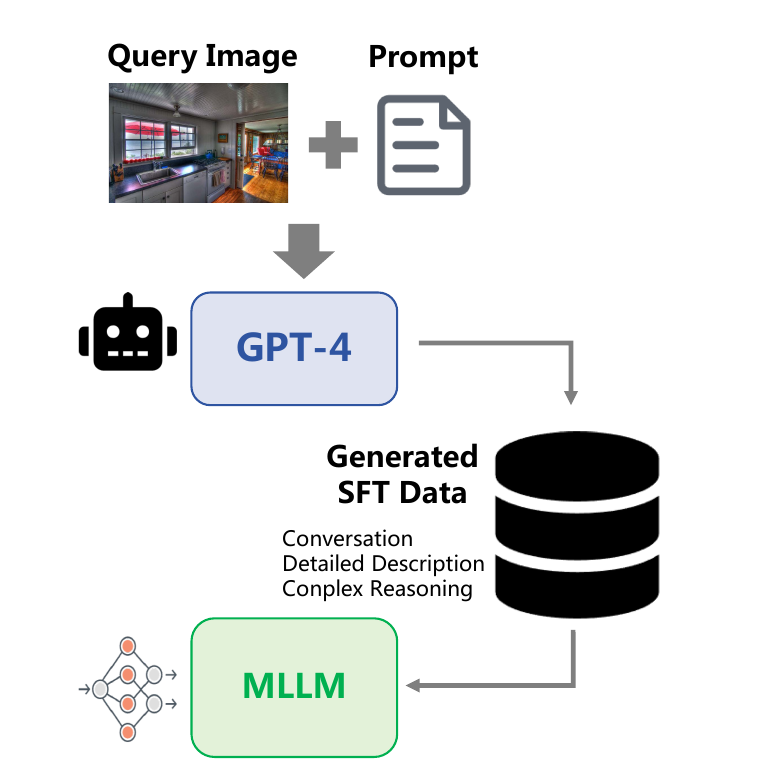}}
    \subfloat[Our Method]{\includegraphics[scale=0.35]{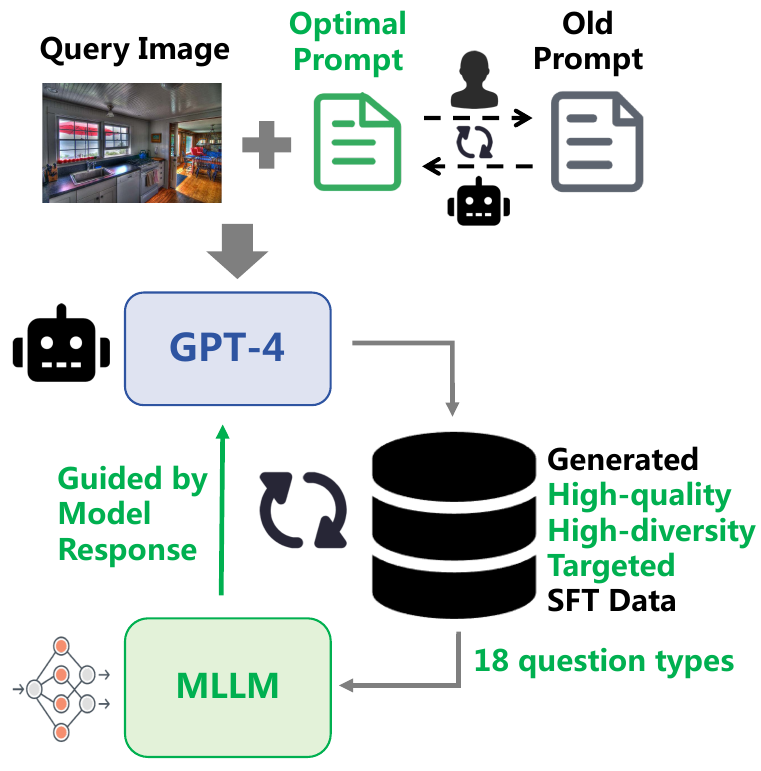}}
    \caption{Comparison of existing methods and our proposed MLLM-DataEngine. 
    Existing methods follow a straightforward one-pass generation process, which separates data generation and evaluation and is unable to achieve targeted and effective model improvement. In contrast, our proposed MLLM-DataEngine is a closed-loop of generation-training-evaluation-generation which lead to targeted model improvement.}
    \label{fig:fig1}
\end{figure*}

\section{Introduction}

The thriving field of Multimodal Large Language Models (MLLM) has seen significant advancements through amalgamating high-quality large visual models \cite{DBLP:conf/iclr/DosovitskiyB0WZ21} and large language models \cite{DBLP:journals/corr/abs-2302-13971, vicuna2023, DBLP:journals/corr/abs-2307-09288}. This integration has given rise to a plenty of sophisticated image-text large models \cite{DBLP:journals/corr/abs-2304-10592, DBLP:journals/corr/abs-2304-08485, DBLP:journals/corr/abs-2305-06500, DBLP:journals/corr/abs-2306-14824, DBLP:conf/nips/AlayracDLMBHLMM22}. To equip these models with capabilities such as image-text dialogue and image content understanding, a two-stage fine-tuning process is typically employed. The first stage aligns image-text features, while the second, crucial to model performance, utilizes high-quality annotated data for instruction fine-tuning. Among these two components, the second stage is pivotal in guaranteeing exceptional model performance. Therefore, obtaining high-quality instruct tuning data is vital for outstanding model performance.

To further explore the capability of MLLMs, there are several efforts on both instruct tuning data collection and benchmark building. For pursuing high-quality multimodel instruct tuning data, recent studies have begun to explore high-quality data generation. Several efforts \cite{DBLP:journals/corr/abs-2305-06500, DBLP:journals/corr/abs-2306-05425} have been undertaken to hand-craftly construct data from public datasets. In contrast, some recent advancements like ChatCaptioner \cite{DBLP:journals/corr/abs-2303-06594} and IdeaGPT \cite{DBLP:journals/corr/abs-2305-14985} using Large Language Models (LLM) like ChatGPT \cite{DBLP:journals/corr/abs-2303-08774} and a Vision-Language Model (VLM) for creating Caption data. LLaVA \cite{DBLP:journals/corr/abs-2304-08485} harnesses GPT-4 \cite{DBLP:journals/corr/abs-2303-08774}, a superior text model, with image annotation for multimodal data generation. For benchmark, corresponding benchmarks for MLLM have also been proposed to furnish a more exhaustive and precise evaluation of MLLM, such as MMBenchmark \cite{DBLP:journals/corr/abs-2307-06281} and MME \cite{DBLP:journals/corr/abs-2306-13394}.

Despite their efforts, the training and evaluation of current MLLMs are isolated. The benchmark could point out the strength and weaknesses of the model \cite{DBLP:journals/corr/abs-2304-02643}, while it is non-trivial to use it as guidance to further improve the model, especially when the benchmark is comprehensive and the weakness includes several different aspects. A straightforward solution is to annotate/collect new data by humans, while its cost is quite large, especially when the models and benchmarks are updated iteratively.

To solve this problem, we propose \textbf{MLLM-DataEngine} to bridge data generation, model training, and evaluation. Our method differs from previous approaches by introducing a closed-loop cycle of generation-training-evaluation-generation where results from the evaluation phase are harnessed to guide the data generation process. Updating MLLM in the loop can generate targeted training data of higher quality and correctness. Specifically: (1) For targeting, we build a bad case pool for the model in each iteration, and propose an Adaptive Bad-case Sampling module to select proper query images and in-context learning examples from the bad case pool, which is essential as feedback to guide further data generation; (2) For quality, we utilize GPT-4 to generated diverse and accurate data for various question type; (3) For correctness, we propose an Interactive Prompt Optimization (IPO) strategy. By multi-round communication between human and GPT-4 on how to avoid generating specific failure cases, the prompt can be effectively optimized. Our contributions are as follows:
\begin{itemize}
  \item We present MLLM-DataEngine, a multimodal engine that fosters a closed loop for data generation, model training, and system evaluation, thus facilitating iterative improvement of model performance and data quality.
  
  \item MLLM-DataEngine guides data generation using model response for targeted model enhancement. Concurrently, we introduce the Adaptive Bad-case Sampling (ABS) module to adaptively choose relevant in-context examples, ensuring high-quality generated data. 
  
  \item In data generation phase of MLLM-DataEngine, we introduce the Interactive Prompt Optimization (IPO) module. Combined with human interaction, it iteratively corrects prompt misunderstandings, yielding substantial improvements in data quality.
  
  \item To validate MLLM-DataEngine's effectiveness, we performed extensive experiments using robust evaluation benchmarks like MMBenchmark and classic VQA datasets like A-OKVQA. Results confirm MLLM-DataEngine's ability to iteratively enhance model performance and data quality.
\end{itemize}

\section{Related Work}

\subsubsection{Multimodel Large Language Model (MLLM)}

With the advancement of transformer architecture \cite{DBLP:conf/nips/VaswaniSPUJGKP17}, Vision-Language models (VLMs) like CLIP \cite{DBLP:conf/icml/RadfordKHRGASAM21}, BLIP \cite{DBLP:conf/icml/0001LXH22}, and BEiT \cite{DBLP:conf/iclr/Bao0PW22} have notably evolved. While current VLMs handle a limited range of simple tasks, the introduction of powerful language models like FLAN-T5 \cite{DBLP:journals/corr/abs-2210-11416}, LLAMA \cite{DBLP:journals/corr/abs-2302-13971}, and Chat-GPT/GPT-4 \cite{DBLP:journals/corr/abs-2303-08774} has considerably expanded their performance and capabilities. BEiT-3 \cite{DBLP:journals/corr/abs-2208-10442} notably performs in multiple vision-language tasks by considering visual features as a unique foreign language. Similarly, LLAVA \cite{DBLP:journals/corr/abs-2304-08485}, MiniGPT-4 \cite{DBLP:journals/corr/abs-2304-10592}, BLIP-2 \cite{DBLP:journals/corr/abs-2301-12597}, and InstructBLIP \cite{DBLP:journals/corr/abs-2305-06500} use a simplified architecture that leverages ViT \cite{DBLP:conf/iclr/DosovitskiyB0WZ21} and QFormer \cite{DBLP:journals/corr/abs-2301-12597} for image feature extraction and an FC layer for aligning these features into language model space. We use this simple architecture in our experiments because it is highly representative and won't damage the ability of original LLM.

\begin{figure*}[ht]
    \centering
    \includegraphics[width=\textwidth]{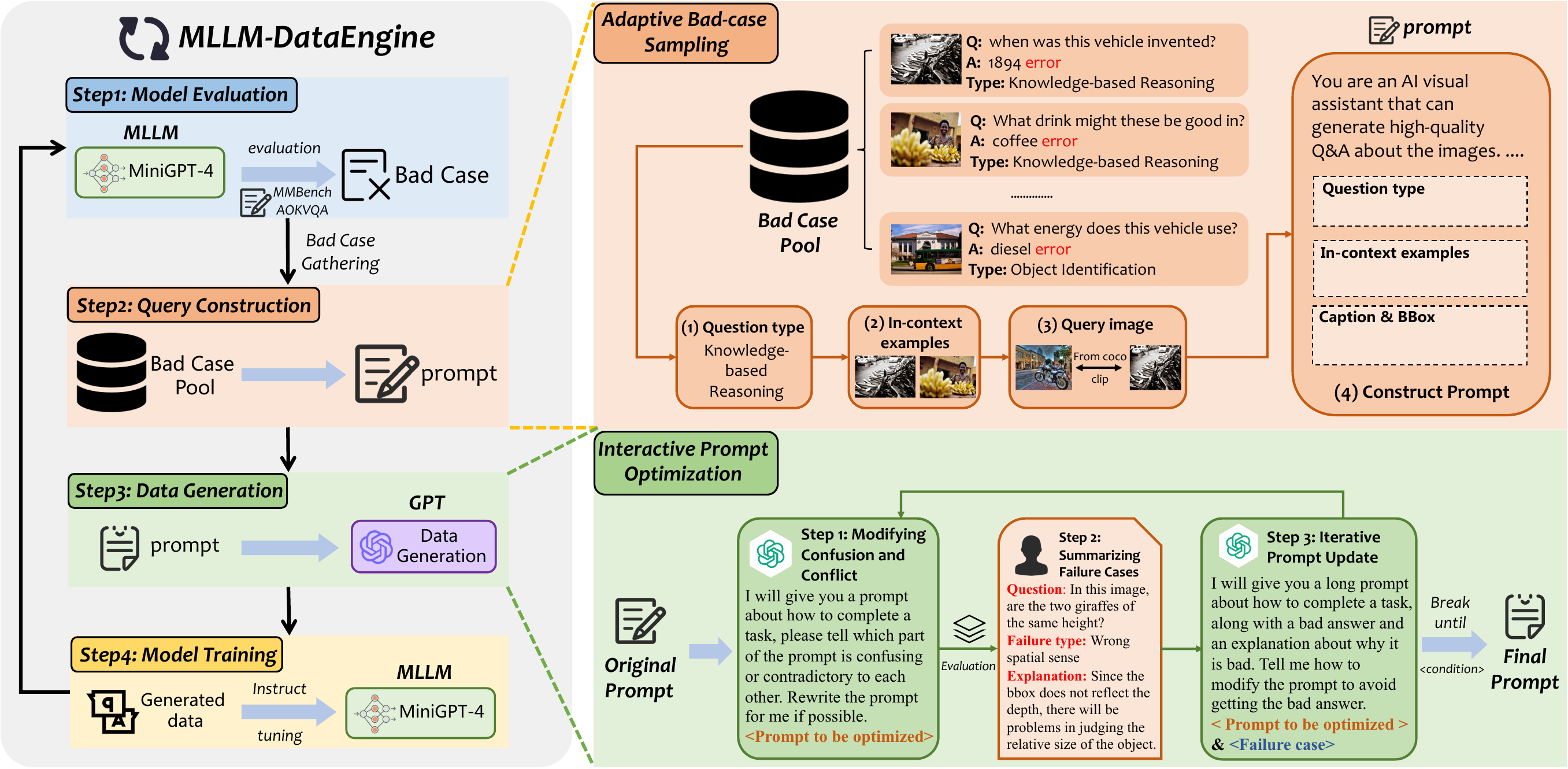}
    \caption{Our proposed MLLM-DataEngine. The whole process is divided into 4 steps. (1) \textbf{Model Evaluation.} We first test the base model on the public benchmark to get the bad cases and classify the bad cases to get the bad case pool. (2) \textbf{Query Construction.} After bad cases are obtained, Adaptive Bad-case sampling (ABS) is proposed to select queries for further data generation. (3) \textbf{Data Generation.} Selected queries are fed to GPT-4 to generate data. Interactive Prompt Optimization (IPO) is proposed to ensure the high quality of generated data. (4) \textbf{Model Training.} Model is fine-tuned on the latest generated data and loops back to the beginning of data engine. }
    \label{fig:fig2}
\end{figure*}

\subsubsection{Multi-modal Instruction Tunning Data}
In computer vision, research increasingly utilizes advanced language models like LLaMa \cite{DBLP:journals/corr/abs-2302-13971}, Vicuna \cite{vicuna2023}, and Flan-T5 \cite{DBLP:journals/corr/abs-2210-11416} to build robust multimodal models such as MiniGPT-4 \cite{DBLP:journals/corr/abs-2304-10592}, LLaVA \cite{DBLP:journals/corr/abs-2304-08485}, and InstructBLIP \cite{DBLP:journals/corr/abs-2305-06500}. These methods employ a two-stage training paradigm, primarily using pre-training data from LAION \cite{DBLP:conf/nips/SchuhmannBVGWCC22} and CC \cite{sharma2018conceptual, changpinyo2021cc12m}, followed by multimodal instruction fine-tuning.

Creating multimodal instruction datasets involves deep image understanding and text development. MiniGPT-4 \cite{DBLP:journals/corr/abs-2304-10592} uses a feature-aligned model for CC dataset interpretation and ChatGPT for initial filtering, curating 3,500 image-text pairs for refinement. LLaVA \cite{DBLP:journals/corr/abs-2304-08485} generates data via language-only GPT-4, which restricts scalability due to manual annotation. To broaden task diversity, InstructBLIP \cite{DBLP:journals/corr/abs-2305-06500} introduces an instruction template construction strategy, converting 26 datasets into fine-tuning data. Meanwhile, MIMIC \cite{DBLP:journals/corr/abs-2305-03726, DBLP:journals/corr/abs-2306-05425} compiles larger-scale instruction fine-tuning datasets. These datasets need human annotation and are limited in both specificity and accuracy. This paper introduces a self-guided, model-driven method for generating high-quality fine-tuning data for any new image.

\begin{figure*}[t]
    \centering
    \includegraphics[width=1.0\textwidth]{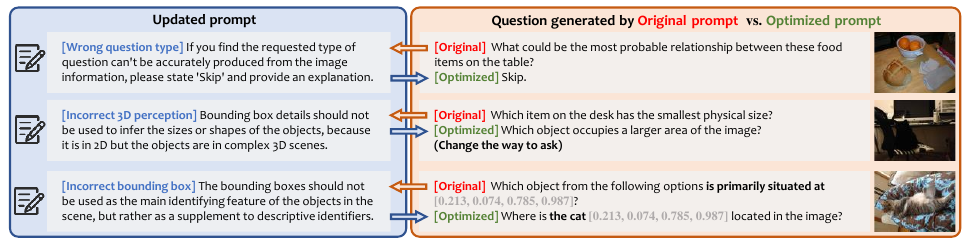}
    \caption{Our proposed Interactive Prompt Optimization (IPO) has introduced key enhancements, primarily: (1) Advanced adherence to the question type, while directly neglecting unsuitable ones; (2) Noteworthy advancements in handling spatial and imaginative questions; (3) Avoided incorrect use of the bounding box.}
    \label{fig:Progress_on_prompt_iteration}
\end{figure*}

\subsubsection{Evaluation of Multi-model Large Language Models}

MLLM's capability dimensions are so vast that they can't be accurately represented by a single or few traditional vision-language tasks. InstructBLIP conducts evaluations on many downstream tasks and datasets (primarily VQA datasets such as A-OKVQA \cite{DBLP:conf/eccv/SchwenkKCMM22}, OKVQA \cite{DBLP:conf/cvpr/MarinoRFM19}, ScienceQA \cite{DBLP:conf/nips/LuMX0CZTCK22} and etc.). Fully evaluating models using public datasets is insufficient, hence the proposal for more exhaustive, objective evaluations suitable for MLLM. LLAVA evaluates models based on 90 open-ended queries and leaves judgment to GPT-4. MMBenchmark \cite{DBLP:journals/corr/abs-2307-06281} and MME \cite{DBLP:journals/corr/abs-2306-13394} propose a benchmark comprising thousands of questions on multiple MLLM ability dimensions, in order to provide a comprehensive evaluation. Except for evaluation, segment Anything \cite{DBLP:journals/corr/abs-2304-02643} shows that the bad case during evaluation can be advantageous to the data refinement process. Inspired by this, we use bad cases during evaluation to guide the data generation process.

\section{Our Approach}

The framework of MLLM-DataEngine showcased in this paper establishes a cyclical process for iterative enhancement between the model and data. The system harnesses generated data to improve the ability of the model, while feedback from model is employed inversely to steer data generation. This approach allows simultaneous optimization of model and data across multiple iterations, facilitating quick training of high-performance models. 

As Figure~\ref{fig:fig2} illustrates, each iteration involves four steps: (1) Model Evaluation. The model's capabilities are systematically evaluated across various dimensions. Then its weaknesses are identified, and bad cases are collected. (2) Query Construction. In the context of these bad cases, we propose \textbf{Adaptive Bad-case Sampling (ABS)} to construct queries for future data generation. (3) Data Generation. Constructed queries are integrated into Prompt, guiding GPT-4 to generate data with the feedback. Moreover, \textbf{Interactive Prompt Optimization (IPO)} is proposed to enhance the instruct-following ability of GPT-4. (4) Model Training. The model parameters are fine-tuned according to the latest generated data. Then MLLM-DataEngine loops back to Model Evaluation for new model performance testing.

\subsection{Model Evaluation}

In this step, we evaluate the model's performance to identify the weakness of the model, and use bad cases as feedback to guide further data generation. In this paper, we utilize the open-source, instruction-fine-tuned MiniGPT-4 \cite{DBLP:journals/corr/abs-2304-10592} as the initial model for the first round of model evaluation. Given that traditional single-task evaluation methods, such as VQA and Caption, cannot comprehensively and accurately assess the capabilities of MLLM, we have chosen MMBenchmark, which involves over 20 ability dimensions, ranging from simple image perception to complex future predictions, as our evaluation standard. Moreover, to validate the proficiency of our proposed method in specific domains, we also evaluate it on the A-OKVQA dataset, one of the most challenging and high-quality mainstream VQA datasets. After evaluation, the bad cases are collected, and classified into 18 question types using GPT-4 (refer to supplemental materials for detail) to build the bad case pool for the model.

\subsection{Query Construction}
After the bad case pool is established, we construct queries for further data generation, which are composed of three key components: (1) An effective selection of proper query images and in-context learning examples, and (2) detailed image information described in language (provided by COCO annotation), (3) A clear, concise prompt for GPT (which will be elaborated on following section). Each of these three components plays a crucial role in the generation of high-quality data.

To ensure effective selection of proper query images and in-context learning examples, we introduce Adaptive Bad-case Sampling (ABS). It comprises three steps: (1) A question type is selected randomly. The probability of selection is inversely proportional to the scores achieved by the model in various ability dimensions. This approach provides guidance for supplementing the data needed to address the model's weakness. (2) Two bad cases of this question type are chosen randomly from the bad case pool to serve as in-context examples, assisting GPT-4 in understanding the required question type effectively and guiding GPT-4 to generate valuable data in following data generation phase. (3) Query images are chosen from the COCO dataset. Among them, half of the images are randomly selected; the remaining half use CLIP to match an image similar to the image of in-context example. This method can guarantee the diversity of generated data while also ensuring query images are suitable for generating QA with the selected question type. All query images are sourced exclusively from the COCO dataset, which includes image captions, object category information, and location details. After the preparation of query images, question types, and in-context examples, the way for future data generation is paved.

\begin{table*}[ht]
    \centering
    \scalebox{0.86}{
    \begin{tabular}{l|cc|c|cc|cc}
        \toprule
        \multirow{2}{*}{\textbf{Base SFT Data}} & \multicolumn{2}{c|}{\textbf{GPTVQA}} &
        \multirow{2}{*}{\textbf{LLM}} & \multicolumn{2}{c|}{\textbf{QMAE}} & \multicolumn{2}{c}{\textbf{QMA}} \\
        & \textbf{Round1} & \textbf{Round2} & & \textbf{fc} & \textbf{lora} & \textbf{fc} & \textbf{lora} \\
        \midrule
        \multirow{6}{*}{\textbf{CCSBUAlign}} & & & \multirow{3}{*}{Vicuna-7B} & 24.3 & -  & - & -  \\
         & \checkmark & & & 26.6 & 38.9 & 30.5 & 40.8  \\
         & \checkmark & \checkmark & & \textbf{27.6} & \textbf{41.9} & \textbf{35.2} & \textbf{50.2} \\
        \cmidrule{2-8}
         & & & \multirow{3}{*}{Vicuna-13B} & 42.3 & - & - & - \\
         & \checkmark & &  & 42.7 & 32.6 & 42.1 & 48.8\\
         & \checkmark & \checkmark &  & \textbf{43.5} & \textbf{44.4} & \textbf{44.0} & \textbf{49.4} \\
        \midrule
        \multirow{6}{*}{\textbf{CCSBUAlign + A-OKVQA}} & & & \multirow{3}{*}{Vicuna-7B} & 25.7 & 25.7 & 38.5 & 45.8 \\
         & \checkmark & &  & 26.4 & 40.6 & 38.4 & 47.1 \\
         & \checkmark & \checkmark &  & \textbf{26.5} & \textbf{46.7} & \textbf{39.2} & \textbf{52.7} \\
        \cmidrule{2-8}
         & & & \multirow{3}{*}{Vicuna-13B} & 44.4 & 46.1 & 45.8 & 52.6 \\
         & \checkmark & &  & \textbf{45.7} & 48.1 & 46.1 & 52.5 \\
         & \checkmark & \checkmark &  & 44.6 & \textbf{49.2} & \textbf{46.5} & \textbf{56.1} \\
        \midrule
        \multirow{3}{*}{\textbf{CCSBUAlign + LLAVA150K}} & & & \multirow{3}{*}{Vicuna-7B} & \textbf{31.0} & 26.0 & - & - \\
         & \checkmark & &  & 28.2 & 36.9 & 31.6 & 44.4 \\
         & \checkmark & \checkmark & & 26.1 & \textbf{38.0} & \textbf{36.3} & \textbf{47.8} \\
        \bottomrule
    \end{tabular}}
    \caption{Experiments on MMBenchmark dev. \emph{GPTVQA} refer to the QA data generated by MLLM-DataEngine. \emph{Round1} and \emph{Round2} refers to the data generated in the first and second round of MLLM-DataEngine. \emph{fc} refers to only fine-tuning the linear projector layer of MiniGPT-4. \emph{lora} refers to LORA finetune. The blank cells in the table are due to default settings and do not require modification or additional experiments.}
    \label{table1}
\end{table*}

\subsection{Data Generation}

In this section, we aim to generate QA data with GPT-4 using the previously constructed query. A well-crafted prompt is key for eliciting high-quality responses from GPT, allowing us to surpass the LLaVA dataset by diversifying question types and reducing GPT-4-induced illusions. Our prompt should incorporate two elements: (1) The concept of question types, and (2) Object-specific BBox references to minimize illusions \cite{DBLP:journals/corr/abs-2306-15195}. However, GPT-4 tends to falter with complex prompts, resulting in issues such as inappropriate question types, incorrect 3D perception, and faulty bounding boxes (refer to supplementary materials for more details).

To counter this, we introduce \textbf{Interactive Prompt Optimization (IPO)}, illustrated in Figure \ref{fig:fig2}, to tackle complex prompt construction. IPO enhances GPT-4's instruction-following capabilities through a semi-automatic optimization process that fosters human-GPT-4 collaboration. This process unfolds in three phases:

\textbf{Step 1: Eliminating Confusion and Conflict:} We first formulate an initial rule-based manually designed Prompt, leveraging LLaVA's methodology. The Prompt instructs GPT-4 to generate specific QA pairs for given images, including bounding box annotations for each object to reduce illusions. GPT-4 then reviews the current Prompt, pinpointing any confusing or conflicting expressions, allowing us to refine it accordingly.

\textbf{Step 2: Summarizing Failure Cases}: Utilizing the refined Prompt, GPT-4 generates a small QA data batch for testing. We manually classify failure cases by type, providing an explanation detailing why it is considered as failure case.

\textbf{Step 3: Prompt Correction based on Failure Cases:} GPT-4 analyses the causes of unreasonable data generation and suggests prompt modifications, using the current template, few-shot failure examples, and their explanations as inputs. The Prompt is optimized as per these suggestions and returns to Step 1 for conflict checks.

The final Prompt is finalized when failure cases drop to an acceptable threshold. It is then employed to generate high-quality QA data with the constructed query. As Figure \ref{fig:Progress_on_prompt_iteration} demonstrates, our approach enhances: (1) Adherence to the question type, skipping unsuitable ones; (2) Noteworthy advancements in handling spatial and imaginative questions; (3) Accuracy in bounding box usage. Refer to supplemental materials for the specific Prompt discussed here. By integrating this optimized prompt with our queries, we can now guide GPT-4 to generate a suitable incremental dataset for training.

\subsection{Model Training}

Upon generating the latest data, we harness all previously generated data from preceding cycles to fine-tune our model. To thoroughly explore the impact of generated data on model performance, we employ two mainstream model fine-tuning strategies:

\textbf{Fine-Tuning Only the Projector Layer (\emph{onlyfc}).} This strategy only updates a few parameters between the vision model and the language model, resulting in the Language-Linked Model (LLM) learning less from the fine-tuning process, but it is highly efficient.

\textbf{LoRA: Low-Rank Adaptation of Large Language Models (\emph{lora})}. As per this strategy, which has been outlined by \cite{DBLP:conf/iclr/HuSWALWWC22}, a low-rank adaptation matrix is employed to fine-tune the LLM. It offers a strategic balance between the cost and effectiveness of fine-tuning. Utilizing \emph{lora} allows the LLM to gain enhanced knowledge from the fine-tuning process while concurrently conserving computational resources.

After the model is fine-tuned on the latest data, the process loops back to the beginning. The newly fine-tuned model is evaluated to obtain the new round of bad cases. Consequently, a new cycle of the data engine is initiated.

\section{Experiments}

We experimentally validate our proposed MLLM-DataEngine's data quality and iterative refinement of both data and model in MLLM-DataEngine. We first test the data quality on A-OKVQA and MMBenchmark, then evaluate on MME \cite{DBLP:journals/corr/abs-2306-13394}. Next, quantitative studies affirm IPO's efficacy in enhancing data quality and diversity. Further, ablation studies reveal the effectiveness of ABS and IPO in MLLM-DataEngine. Finally, we show that MLLM-DataEngine's closed-loop refinement yields higher-quality data than a simple one-pass strategy.

\subsection{Implementation Details}

MiniGPT-4 is chosen for our MLLM architecture, which utilizes BLIP-2 \cite{DBLP:journals/corr/abs-2301-12597}, a ViT-G/14 and Q-former amalgamation, as the visual feature extractor. These visual features are fed directly into the LLM via a linear projector, acting as a soft prompt. The language model component utilizes Vicuna-7B and Vicuna-13B.

The MLLM-DataEngine creates a dataset termed \textbf{GPTVQA}, comprising Question-Answer (Q-A) pairs that include question (Q), multiple choices (M), answer (A), and rationale (E). Two formats, QMAE and QMA, are mainly explored in our experiments. We run the data engine for two rounds and generate 5K and 18K data respectively.

\begin{table}[t]
    \centering
    \small
    \scalebox{0.95}{
    \begin{tabular}{l|cc|c|cc|cc}
        \toprule
        \multirow{2}{*}{\textbf{SFT Data}} & \multirow{2}{*}{\textbf{R1}} & \multirow{2}{*}{\textbf{R2}} &
        \multirow{2}{*}{\textbf{LLM}} & \multicolumn{2}{c|}{\textbf{QMAE}} & \multicolumn{2}{c}{\textbf{QMA}} \\
        & & & & \textbf{MC} & \textbf{DA} & \textbf{MC} & \textbf{DA} \\
        \midrule
        \multirow{6}{*}{\makecell[l]{A-OKVQA}} & & & 7B & 70.1 & 59.1 & 70.2 & 59.1 \\
         & \checkmark & & 7B & 71.0 & 60.3 & 71.8 & 60.8 \\
         & \checkmark & \checkmark & 7B & \textbf{72.1} & \textbf{61.0} & \textbf{73.6} & \textbf{62.0} \\
         \cmidrule{2-8}
        & & & 13B & 73.1 & 62.1 & 74.8 & 62.6 \\
         & \checkmark & & 13B & \textbf{74.5} & \textbf{62.4} & 74.7 & 63.1 \\
         & \checkmark & \checkmark & 13B & 74.0 & 61.9 & \textbf{75.5} & \textbf{63.3} \\
        \bottomrule
    \end{tabular}}
    \caption{Experiments on A-OKVQA val split.}
    \label{table2}
\end{table}

\begin{table}[t]
    \centering
    \begin{tabular}{c|cc|cc|cc}
        \toprule
        \multirow{2}{*}{\textbf{\makecell[c]{LLM}}} & \multirow{2}{*}{\textbf{R1}} & \multirow{2}{*}{\textbf{R2}} &
        \multicolumn{2}{c|}{\textbf{QMAE}} & \multicolumn{2}{c}{\textbf{QMA}} \\
        & & & \textbf{fc} & \textbf{lora} &   \textbf{fc} &   \textbf{lora} \\
        \midrule
        \multirow{3}{*}{{\makecell[l]{7B}}} & & &  23.0 & -  & - & -  \\
        & \checkmark & &   \textbf{26.6} &   \textbf{38.7} &   32.8 &   42.3 \\
        & \checkmark & \checkmark & 25.5 & 37.6 & \textbf{37.2} & \textbf{48.6}\\
        \midrule
        \multirow{3}{*}{{\makecell[l]{13B}}} & & & 42.3 & - &  - & -  \\
        & \checkmark & &   42.8 & 36.7 & 43.2 &   48.2 \\
        & \checkmark & \checkmark & \textbf{42.9} & \textbf{44.8} & \textbf{44.5} & \textbf{48.3} \\
        \bottomrule
    \end{tabular}
    \caption{Experiments on MMBenchmark test.}
    \label{table3}
\end{table}

Experiments involve variable settings including fine-tuning the linear projector layer (\emph{fc}) and applying LORA fine-tuning \cite{DBLP:conf/iclr/HuSWALWWC22} (\emph{lora}). We use 4 A100-80G GPUs and batch size is set to 64. We employ AdamW optimizer with $\beta_1=0.9$, $\beta_2=0.999$, and weight decay of 0.05. The learning rate warms up during the first 200 steps, starting from $1e^{-6}$ and reaches peak values for \emph{fc}, \emph{lora} at  $3e^{-5}$, $3e^{-4}$, respectively. Cosine learning rate scheduler is employed. For \emph{lora} fine-tune, we fine-tune $q$ and $k$ in the attention layer. Rank $r$ in \emph{lora} is set to 8. If not specified, we instruct tune the model by 10 epochs, 300 iterations per epoch. During instruct tuning, we use A-OKVQA as the validation set and choose the best model in the final evaluation. 

\begin{table}[t]
    \centering
    \footnotesize
    \begin{tabular}{c|cc|cc|cc}
        \toprule
        \multirow{2}{*}{\textbf{{SFT Data}}} & \multirow{2}{*}{\textbf{R1}} & \multirow{2}{*}{\textbf{R2}} &
        \multicolumn{2}{c|}{\textbf{Perception}} & \multicolumn{2}{c}{\textbf{Cognition}} \\
        & & & \textbf{fc} & \textbf{lora} & \textbf{fc} & \textbf{lora} \\
        \midrule
        \multirow{3}{*}{{{CCSBUAlign}}} & & &  670.1 & - & 125.0 & - \\
        & \checkmark & &   714.2 &   708.9 & 115.3 & 206.4 \\
        & \checkmark & \checkmark & \textbf{732.5} & \textbf{742.5} & \textbf{148.2} & \textbf{208.9} \\
        \bottomrule
    \end{tabular}
    \caption{Experiments on MME using Vicuna-7B. Perception and Cognition are two main abilities tested in MME, which is measured on 14 subtasks.}
    \label{table4}
\end{table}

\subsection{Experiments on MMBenchmark}

Three types of data, namely (1) CCSBUAlign, (2) CCSBUAlign and A-OKVQA, (3) CCSBUAlign and LLAVA150K \cite{DBLP:journals/corr/abs-2304-08485}, are employed to construct baseline models. GPTVQA is then incorporated for validation purposes. As shown in Table \ref{table1}, results on MMBenchmark dev reflect the continuous improvements from using GPTVQA generated by MLLM-DataEngine across different formats (QMA and QMAE) and settings (\emph{onlyfc} and \emph{lora}).

Further experiments using a stronger baseline model (CCSBUAlign + A-OKVQA) yield scores between 40 and 50 on MMBenchmark, indicating that A-OKVQA serves as a solid comparative dataset. Notably, even with this robust model, GPTVQA contributes to a near 5\% improvement. Similar results are observed with another stronger baseline model (CCSBUAlign + LLAVA150K).

Intriguingly, we find that the QMA format outperforms QMAE in both \emph{onlyfc} and \emph{lora} settings, possibly due to LLM's preference for shorter token length when partially open. Furthermore, performance variation is noted across settings (\emph{fc} and \emph{lora}). The improvement lead by \emph{fc} is limited, while \emph{lora} shows significant improvement and achieves the highest performance.

\begin{table}[t]
    \large
    \centering
    \begin{tabular}{lcc}
        \toprule
        \textbf{Failure type} & \textbf{Before} & \textbf{After}\\
        \midrule
          Incorrect bounding box & 24 & 0\\
          Illusion & 20 & 2\\
          Incorrect 3D perception & 15 & 3\\  
          Wrong question type & 8 & 0\\
          Illogical question & 8 & 0\\
        \bottomrule
    \end{tabular}
    \caption{The failure cases statistic before and after IPO. }
    \label{table: failure_case_statistic}
\end{table}

\begin{table}[t]
    \centering
    \scalebox{0.92}{
    \begin{tabular}{l|cc}
        \toprule
        \textbf{Model} & \textbf{A-OKVQA} & \textbf{GPTVQA (r1+r2)}\\
        \midrule
          Instance Num. & 17056 & 23164\\
          Unique Q & 16192 (94.9\%) & 18936 (81.7\%)\\
          Unique A & 16983 (99.5\%) & 23163 (100\%)\\
          Avg. length (Q/A) & 8.81/16.38 & 12.01/63.89\\
          Unique noun (A) & 4501 & 5296\\
          Mean Q distance (Q) & 0.897 & 0.814\\
        \bottomrule
    \end{tabular}}
    \caption{Comparison of data diversity between A-OKVQA and GPTVQA. Instance Num. is the total number of QA instances. Q and A represent question and answer, respectively.}
    \label{table: data_diversity}
\end{table}

\begin{figure*}[t]
    \centering
    \includegraphics[width=\textwidth]{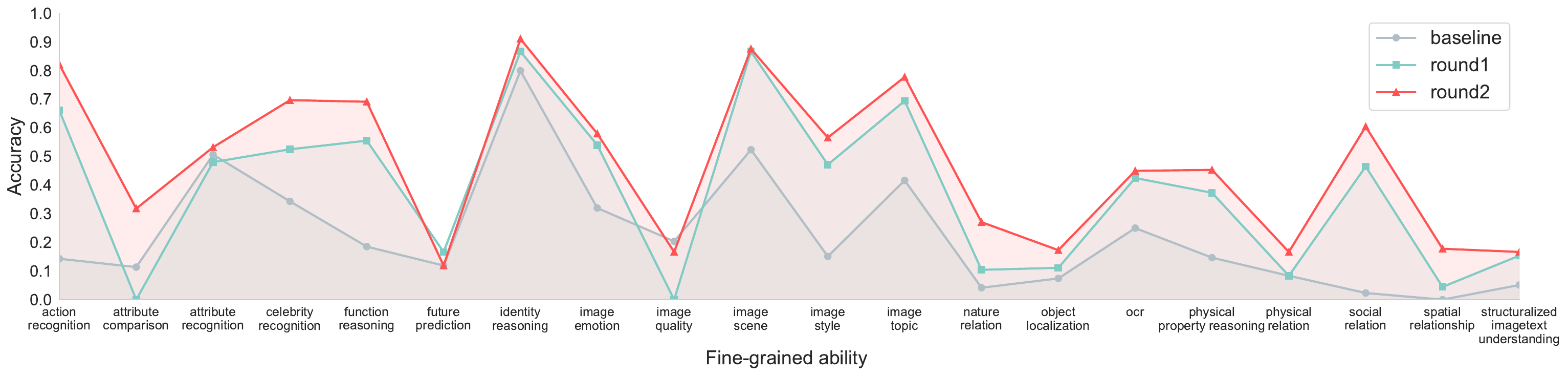}
    \caption{Fine-grained ability comparison of each round model.}
    \label{fig:ability}
\end{figure*}

\subsection{Experiments on A-OKVQA}

We further verify the quality of GPTVQA by conducting comprehensive experiments on traditional visual question-answering dataset. The A-OKVQA validation split is specifically chosen for experiments. We evaluate our results using Multi-Choice (MC) and Direct Answer (DA), which are standard metrics utilized by A-OKVQA. A rank-based method is implemented to predict answers, with the results displayed in Table \ref{table2}. Results suggest that GPTVQA significantly improves accuracy on the A-OKVQA task. This enhancement is consistently observed across various settings and input formats, thus further validating the outstanding quality of GPTVQA.

\subsection{Experiments on additional benchmarks}

We further validate the effectiveness of MLLM-DataEngine and quality of GPTVQA via additional benchmark experiments. An initial evaluation on MMBenchmark test displays consistent model improvement during each refinement round. Results are shown in Table \ref{table3}. Further evaluations on another open-ended benchmark, MME \cite{DBLP:journals/corr/abs-2306-13394}, used for MLLM evaluation, also show enhanced performance from the GPTVQA in Table \ref{table4}. These supplementary experiments affirm the efficacy of both model and data refinement across various benchmarks, thereby confirming that our proposed MLLM-DataEngine does not overfit specific evaluation benchmarks.

\subsection{Experiments on data quality and diversity}

We quantitatively assess IPO by comparing failure cases from initial and optimized Prompts. Table \ref{table: failure_case_statistic} shows that only five out of 77 errors persist with the optimized Prompt. An additional 100 questions are tested; eight have incorrect answers due to limitations in COCO annotation or GPT-4's understanding of illusions and 3D scenarios (see supplementary). These can potentially be mitigated by providing more image information and enhancing the vision-language model. Besides, as shown in Table \ref{table: data_diversity}, we compare dataset diversity between A-OKVQA and GPTVQA. Results show that our generated data achieves almost fair performance in data diversity. Furthermore, our data has a significantly higher number of unique nouns and average answer length than A-OKVQA.

\begin{table}[t]
    \centering
    \begin{tabular}{c|c|ccc}
        \toprule
        \multirow{2}{*}{\textbf{prompt}} & 
        \multirow{2}{*}{\textbf{MMBench Dev}} & \multicolumn{2}{c}{\textbf{A-OKVQA val}} \\
        & & \textbf{MC} & \textbf{DA} \\
        \midrule
        original & 37.8 & 54.8 & 47.4 \\
        optimal & 42.5 \color{red}{(+4.7)} & 60.0 \color{red}{(+5.2)} & 50.4 \color{red}{(+3.0)} \\
        \bottomrule
    \end{tabular}
    \caption{Ablation studies on Interactive Prompt Optimization (IPO).}
    \label{table5}
\end{table}

\begin{table}[t]
    \centering
    \scalebox{0.95}{
    \begin{tabular}{c|c|ccc}
        \toprule
        \multirow{2}{*}{\textbf{SFT Data}} & 
        \multirow{2}{*}{\textbf{MMBench}} & \multicolumn{2}{c}{\textbf{A-OKVQA}} \\
        & & \textbf{MC} & \textbf{DA} \\
        \midrule
        Round1 2.5K + Round1 2.5K & 40.8 & 55.7 & 47.5 \\
        Round1 2.5K + \color{red}{Round2 2.5K} &  46.0 & 62.2 &  52.1 \\
        \bottomrule
    \end{tabular}}
    \caption{Ablation studies on data quality of each round. Results show iterative refinement in our proposed MLLM-DataEngine can generate higher-quality data than direct one-pass generation. }
    \label{table6}
    \vskip -1\baselineskip plus -1fil
\end{table}

\subsection{Ablation Studies}

\subsubsection{Ablation on improvements of fine-grained ability}

We compare the fine-grained abilities among baseline, round 1, and round 2 models, shown in Figure \ref{fig:ability}. Each iteration of our data engine significantly improves most abilities, demonstrating the efficacy of Adaptive Bad-case Sampling (ABS) in selecting proper and diverse in-context examples. This strategy is helpful for creating diverse data, crucial for comprehensive model performance improvements.

\subsubsection{Ablation on Interactive Prompt Optimization (IPO)}

Ablation studies on Interactive Prompt Optimization (IPO) are shown in Table \ref{table5}, where \emph{original} refers to the original prompt, and \emph{optimal} refers to the prompt from interactive optimization iteration (see supplementary for details). Using 1,500 samples for initial and optimally modified prompts, improvements of 3-5\% were observed in both MMBenchmark and A-OKVQA evaluations using the optimal prompt. This suggests an optimized prompt significantly enhances data quality.

\subsubsection{Ablation on data quality refinement}

To validate the effectiveness of round-by-round refinement in improving data quality, as compared to a single-round direct data generation approach, we carry out the following experiments. We randomly sample 5k data points from the data generated in round 2 and used this subset to replace an equivalent quantity of data in round 1. The results of this experiment are detailed in Table \ref{table6}. Our findings indicate that the quality of data from round 2 surpasses that of additional data directly produced in round 1. These results validate that iterative refinement in our proposed MLLM-DataEngine can generate higher-quality data than direct one-pass generation.

\section{Conclusion}
This paper presents MLLM-DataEngine, a framework for generating high-quality, targeted instruction fine-tuning data, addressing model weaknesses and forming a closed training loop for large multi-modal models. This new paradigm enables comprehensive training and quick iteration of these models. During data generation, we introduce Adaptive Bad-case Sampling for appropriate in-context examples selection, and Interactive Prompt Optimization to minimize prompt-induced misunderstanding, thereby securing the generation of high-quality data. We aim to apply this approach to diverse multi-modal tasks across various modalities, anticipating that MLLM-DataEngine will advance multi-modal artificial intelligence research.



\newpage

\bibliography{data_engine}

\begin{thebibliography}{32}
\providecommand{\natexlab}[1]{#1}

\bibitem[{Alayrac et~al.(2022)Alayrac, Donahue, Luc, Miech, Barr, Hasson, Lenc,
  Mensch, Millican, Reynolds, Ring, Rutherford, Cabi, Han, Gong, Samangooei,
  Monteiro, Menick, Borgeaud, Brock, Nematzadeh, Sharifzadeh, Binkowski,
  Barreira, Vinyals, Zisserman, and
  Simonyan}]{DBLP:conf/nips/AlayracDLMBHLMM22}
Alayrac, J.; Donahue, J.; Luc, P.; Miech, A.; Barr, I.; Hasson, Y.; Lenc, K.;
  Mensch, A.; Millican, K.; Reynolds, M.; Ring, R.; Rutherford, E.; Cabi, S.;
  Han, T.; Gong, Z.; Samangooei, S.; Monteiro, M.; Menick, J.~L.; Borgeaud, S.;
  Brock, A.; Nematzadeh, A.; Sharifzadeh, S.; Binkowski, M.; Barreira, R.;
  Vinyals, O.; Zisserman, A.; and Simonyan, K. 2022.
\newblock Flamingo: a Visual Language Model for Few-Shot Learning.
\newblock In \emph{NIPS}.

\bibitem[{Bao et~al.(2022)Bao, Dong, Piao, and Wei}]{DBLP:conf/iclr/Bao0PW22}
Bao, H.; Dong, L.; Piao, S.; and Wei, F. 2022.
\newblock BEiT: {BERT} Pre-Training of Image Transformers.
\newblock In \emph{{ICLR}}.

\bibitem[{Changpinyo et~al.(2021)Changpinyo, Sharma, Ding, and
  Soricut}]{changpinyo2021cc12m}
Changpinyo, S.; Sharma, P.; Ding, N.; and Soricut, R. 2021.
\newblock {Conceptual 12M}: Pushing Web-Scale Image-Text Pre-Training To
  Recognize Long-Tail Visual Concepts.
\newblock In \emph{CVPR}.

\bibitem[{Chen et~al.(2023)Chen, Zhang, Zeng, Zhang, Zhu, and
  Zhao}]{DBLP:journals/corr/abs-2306-15195}
Chen, K.; Zhang, Z.; Zeng, W.; Zhang, R.; Zhu, F.; and Zhao, R. 2023.
\newblock Shikra: Unleashing Multimodal LLM's Referential Dialogue Magic.
\newblock \emph{arXiv}.

\bibitem[{Chiang et~al.(2023)Chiang, Li, Lin, Sheng, Wu, Zhang, Zheng, Zhuang,
  Zhuang, Gonzalez, Stoica, and Xing}]{vicuna2023}
Chiang, W.-L.; Li, Z.; Lin, Z.; Sheng, Y.; Wu, Z.; Zhang, H.; Zheng, L.;
  Zhuang, S.; Zhuang, Y.; Gonzalez, J.~E.; Stoica, I.; and Xing, E.~P. 2023.
\newblock Vicuna: An Open-Source Chatbot Impressing GPT-4 with 90\%* ChatGPT
  Quality.

\bibitem[{Chung et~al.(2022)Chung, Hou, Longpre, Zoph, Tay, Fedus, Li, Wang,
  Dehghani, Brahma, Webson, Gu, Dai, Suzgun, Chen, Chowdhery, Narang, Mishra,
  Yu, Zhao, Huang, Dai, Yu, Petrov, Chi, Dean, Devlin, Roberts, Zhou, Le, and
  Wei}]{DBLP:journals/corr/abs-2210-11416}
Chung, H.~W.; Hou, L.; Longpre, S.; Zoph, B.; Tay, Y.; Fedus, W.; Li, E.; Wang,
  X.; Dehghani, M.; Brahma, S.; Webson, A.; Gu, S.~S.; Dai, Z.; Suzgun, M.;
  Chen, X.; Chowdhery, A.; Narang, S.; Mishra, G.; Yu, A.; Zhao, V.~Y.; Huang,
  Y.; Dai, A.~M.; Yu, H.; Petrov, S.; Chi, E.~H.; Dean, J.; Devlin, J.;
  Roberts, A.; Zhou, D.; Le, Q.~V.; and Wei, J. 2022.
\newblock Scaling Instruction-Finetuned Language Models.
\newblock \emph{arXiv}.

\bibitem[{Dai et~al.(2023)Dai, Li, Li, Tiong, Zhao, Wang, Li, Fung, and
  Hoi}]{DBLP:journals/corr/abs-2305-06500}
Dai, W.; Li, J.; Li, D.; Tiong, A. M.~H.; Zhao, J.; Wang, W.; Li, B.; Fung, P.;
  and Hoi, S. C.~H. 2023.
\newblock InstructBLIP: Towards General-purpose Vision-Language Models with
  Instruction Tuning.
\newblock \emph{arXiv}.

\bibitem[{Dosovitskiy et~al.(2021)Dosovitskiy, Beyer, Kolesnikov, Weissenborn,
  Zhai, Unterthiner, Dehghani, Minderer, Heigold, Gelly, Uszkoreit, and
  Houlsby}]{DBLP:conf/iclr/DosovitskiyB0WZ21}
Dosovitskiy, A.; Beyer, L.; Kolesnikov, A.; Weissenborn, D.; Zhai, X.;
  Unterthiner, T.; Dehghani, M.; Minderer, M.; Heigold, G.; Gelly, S.;
  Uszkoreit, J.; and Houlsby, N. 2021.
\newblock An Image is Worth 16x16 Words: Transformers for Image Recognition at
  Scale.
\newblock In \emph{{ICLR}}.

\bibitem[{Fu et~al.(2023)Fu, Chen, Shen, Qin, Zhang, Lin, Qiu, Lin, Yang,
  Zheng, Li, Sun, and Ji}]{DBLP:journals/corr/abs-2306-13394}
Fu, C.; Chen, P.; Shen, Y.; Qin, Y.; Zhang, M.; Lin, X.; Qiu, Z.; Lin, W.;
  Yang, J.; Zheng, X.; Li, K.; Sun, X.; and Ji, R. 2023.
\newblock {MME:} {A} Comprehensive Evaluation Benchmark for Multimodal Large
  Language Models.
\newblock \emph{arXiv}.

\bibitem[{Hu et~al.(2022)Hu, Shen, Wallis, Allen{-}Zhu, Li, Wang, Wang, and
  Chen}]{DBLP:conf/iclr/HuSWALWWC22}
Hu, E.~J.; Shen, Y.; Wallis, P.; Allen{-}Zhu, Z.; Li, Y.; Wang, S.; Wang, L.;
  and Chen, W. 2022.
\newblock LoRA: Low-Rank Adaptation of Large Language Models.
\newblock In \emph{{ICLR}}.

\bibitem[{Kirillov et~al.(2023)Kirillov, Mintun, Ravi, Mao, Rolland, Gustafson,
  Xiao, Whitehead, Berg, Lo, Doll{\'{a}}r, and
  Girshick}]{DBLP:journals/corr/abs-2304-02643}
Kirillov, A.; Mintun, E.; Ravi, N.; Mao, H.; Rolland, C.; Gustafson, L.; Xiao,
  T.; Whitehead, S.; Berg, A.~C.; Lo, W.; Doll{\'{a}}r, P.; and Girshick, R.~B.
  2023.
\newblock Segment Anything.
\newblock \emph{arXiv}.

\bibitem[{Li et~al.(2023{\natexlab{a}})Li, Zhang, Chen, Wang, Pu, Yang, Li, and
  Liu}]{DBLP:journals/corr/abs-2306-05425}
Li, B.; Zhang, Y.; Chen, L.; Wang, J.; Pu, F.; Yang, J.; Li, C.; and Liu, Z.
  2023{\natexlab{a}}.
\newblock {MIMIC-IT:} Multi-Modal In-Context Instruction Tuning.
\newblock \emph{arXiv}.

\bibitem[{Li et~al.(2023{\natexlab{b}})Li, Zhang, Chen, Wang, Yang, and
  Liu}]{DBLP:journals/corr/abs-2305-03726}
Li, B.; Zhang, Y.; Chen, L.; Wang, J.; Yang, J.; and Liu, Z.
  2023{\natexlab{b}}.
\newblock Otter: {A} Multi-Modal Model with In-Context Instruction Tuning.
\newblock \emph{arXiv}.

\bibitem[{Li et~al.(2023{\natexlab{c}})Li, Li, Savarese, and
  Hoi}]{DBLP:journals/corr/abs-2301-12597}
Li, J.; Li, D.; Savarese, S.; and Hoi, S. C.~H. 2023{\natexlab{c}}.
\newblock {BLIP-2:} Bootstrapping Language-Image Pre-training with Frozen Image
  Encoders and Large Language Models.
\newblock \emph{arXiv}.

\bibitem[{Li et~al.(2022)Li, Li, Xiong, and Hoi}]{DBLP:conf/icml/0001LXH22}
Li, J.; Li, D.; Xiong, C.; and Hoi, S. C.~H. 2022.
\newblock {BLIP:} Bootstrapping Language-Image Pre-training for Unified
  Vision-Language Understanding and Generation.
\newblock In \emph{{ICML}}.

\bibitem[{Liu et~al.(2023{\natexlab{a}})Liu, Li, Wu, and
  Lee}]{DBLP:journals/corr/abs-2304-08485}
Liu, H.; Li, C.; Wu, Q.; and Lee, Y.~J. 2023{\natexlab{a}}.
\newblock Visual Instruction Tuning.
\newblock \emph{arXiv}.

\bibitem[{Liu et~al.(2023{\natexlab{b}})Liu, Duan, Zhang, Li, Zhang, Zhao,
  Yuan, Wang, He, Liu, Chen, and Lin}]{DBLP:journals/corr/abs-2307-06281}
Liu, Y.; Duan, H.; Zhang, Y.; Li, B.; Zhang, S.; Zhao, W.; Yuan, Y.; Wang, J.;
  He, C.; Liu, Z.; Chen, K.; and Lin, D. 2023{\natexlab{b}}.
\newblock MMBench: Is Your Multi-modal Model an All-around Player?
\newblock \emph{arXiv}.

\bibitem[{Lu et~al.(2022)Lu, Mishra, Xia, Qiu, Chang, Zhu, Tafjord, Clark, and
  Kalyan}]{DBLP:conf/nips/LuMX0CZTCK22}
Lu, P.; Mishra, S.; Xia, T.; Qiu, L.; Chang, K.; Zhu, S.; Tafjord, O.; Clark,
  P.; and Kalyan, A. 2022.
\newblock Learn to Explain: Multimodal Reasoning via Thought Chains for Science
  Question Answering.
\newblock In \emph{NeurIPS}.

\bibitem[{Marino et~al.(2019)Marino, Rastegari, Farhadi, and
  Mottaghi}]{DBLP:conf/cvpr/MarinoRFM19}
Marino, K.; Rastegari, M.; Farhadi, A.; and Mottaghi, R. 2019.
\newblock {OK-VQA:} {A} Visual Question Answering Benchmark Requiring External
  Knowledge.
\newblock In \emph{{CVPR}}.

\bibitem[{OpenAI(2023)}]{DBLP:journals/corr/abs-2303-08774}
OpenAI. 2023.
\newblock {GPT-4} Technical Report.
\newblock \emph{arXiv}.

\bibitem[{Peng et~al.(2023)Peng, Wang, Dong, Hao, Huang, Ma, and
  Wei}]{DBLP:journals/corr/abs-2306-14824}
Peng, Z.; Wang, W.; Dong, L.; Hao, Y.; Huang, S.; Ma, S.; and Wei, F. 2023.
\newblock Kosmos-2: Grounding Multimodal Large Language Models to the World.
\newblock \emph{arXiv}.

\bibitem[{Radford et~al.(2021)Radford, Kim, Hallacy, Ramesh, Goh, Agarwal,
  Sastry, Askell, Mishkin, Clark, Krueger, and
  Sutskever}]{DBLP:conf/icml/RadfordKHRGASAM21}
Radford, A.; Kim, J.~W.; Hallacy, C.; Ramesh, A.; Goh, G.; Agarwal, S.; Sastry,
  G.; Askell, A.; Mishkin, P.; Clark, J.; Krueger, G.; and Sutskever, I. 2021.
\newblock Learning Transferable Visual Models From Natural Language
  Supervision.
\newblock In \emph{{ICML}}.

\bibitem[{Schuhmann et~al.(2022)Schuhmann, Beaumont, Vencu, Gordon, Wightman,
  Cherti, Coombes, Katta, Mullis, Wortsman, Schramowski, Kundurthy, Crowson,
  Schmidt, Kaczmarczyk, and Jitsev}]{DBLP:conf/nips/SchuhmannBVGWCC22}
Schuhmann, C.; Beaumont, R.; Vencu, R.; Gordon, C.; Wightman, R.; Cherti, M.;
  Coombes, T.; Katta, A.; Mullis, C.; Wortsman, M.; Schramowski, P.; Kundurthy,
  S.; Crowson, K.; Schmidt, L.; Kaczmarczyk, R.; and Jitsev, J. 2022.
\newblock {LAION-5B:} An open large-scale dataset for training next generation
  image-text models.
\newblock In \emph{NeurIPS}.

\bibitem[{Schwenk et~al.(2022)Schwenk, Khandelwal, Clark, Marino, and
  Mottaghi}]{DBLP:conf/eccv/SchwenkKCMM22}
Schwenk, D.; Khandelwal, A.; Clark, C.; Marino, K.; and Mottaghi, R. 2022.
\newblock {A-OKVQA:} {A} Benchmark for Visual Question Answering Using World
  Knowledge.
\newblock In \emph{ECCV}.

\bibitem[{Sharma et~al.(2018)Sharma, Ding, Goodman, and
  Soricut}]{sharma2018conceptual}
Sharma, P.; Ding, N.; Goodman, S.; and Soricut, R. 2018.
\newblock Conceptual Captions: A Cleaned, Hypernymed, Image Alt-text Dataset
  For Automatic Image Captioning.
\newblock In \emph{ACL}.

\bibitem[{Touvron et~al.(2023{\natexlab{a}})Touvron, Lavril, Izacard, Martinet,
  Lachaux, Lacroix, Rozi{\`{e}}re, Goyal, Hambro, Azhar, Rodriguez, Joulin,
  Grave, and Lample}]{DBLP:journals/corr/abs-2302-13971}
Touvron, H.; Lavril, T.; Izacard, G.; Martinet, X.; Lachaux, M.; Lacroix, T.;
  Rozi{\`{e}}re, B.; Goyal, N.; Hambro, E.; Azhar, F.; Rodriguez, A.; Joulin,
  A.; Grave, E.; and Lample, G. 2023{\natexlab{a}}.
\newblock LLaMA: Open and Efficient Foundation Language Models.
\newblock \emph{arXiv}.

\bibitem[{Touvron et~al.(2023{\natexlab{b}})Touvron, Martin, Stone, Albert,
  Almahairi, Babaei, Bashlykov, Batra, Bhargava, Bhosale, Bikel, Blecher,
  Canton{-}Ferrer, Chen, Cucurull, Esiobu, Fernandes, Fu, Fu, Fuller, Gao,
  Goswami, Goyal, Hartshorn, Hosseini, Hou, Inan, Kardas, Kerkez, Khabsa,
  Kloumann, Korenev, Koura, Lachaux, Lavril, Lee, Liskovich, Lu, Mao, Martinet,
  Mihaylov, Mishra, Molybog, Nie, Poulton, Reizenstein, Rungta, Saladi,
  Schelten, Silva, Smith, Subramanian, Tan, Tang, Taylor, Williams, Kuan, Xu,
  Yan, Zarov, Zhang, Fan, Kambadur, Narang, Rodriguez, Stojnic, Edunov, and
  Scialom}]{DBLP:journals/corr/abs-2307-09288}
Touvron, H.; Martin, L.; Stone, K.; Albert, P.; Almahairi, A.; Babaei, Y.;
  Bashlykov, N.; Batra, S.; Bhargava, P.; Bhosale, S.; Bikel, D.; Blecher, L.;
  Canton{-}Ferrer, C.; Chen, M.; Cucurull, G.; Esiobu, D.; Fernandes, J.; Fu,
  J.; Fu, W.; Fuller, B.; Gao, C.; Goswami, V.; Goyal, N.; Hartshorn, A.;
  Hosseini, S.; Hou, R.; Inan, H.; Kardas, M.; Kerkez, V.; Khabsa, M.;
  Kloumann, I.; Korenev, A.; Koura, P.~S.; Lachaux, M.; Lavril, T.; Lee, J.;
  Liskovich, D.; Lu, Y.; Mao, Y.; Martinet, X.; Mihaylov, T.; Mishra, P.;
  Molybog, I.; Nie, Y.; Poulton, A.; Reizenstein, J.; Rungta, R.; Saladi, K.;
  Schelten, A.; Silva, R.; Smith, E.~M.; Subramanian, R.; Tan, X.~E.; Tang, B.;
  Taylor, R.; Williams, A.; Kuan, J.~X.; Xu, P.; Yan, Z.; Zarov, I.; Zhang, Y.;
  Fan, A.; Kambadur, M.; Narang, S.; Rodriguez, A.; Stojnic, R.; Edunov, S.;
  and Scialom, T. 2023{\natexlab{b}}.
\newblock Llama 2: Open Foundation and Fine-Tuned Chat Models.
\newblock \emph{arXiv}.

\bibitem[{Vaswani et~al.(2017)Vaswani, Shazeer, Parmar, Uszkoreit, Jones,
  Gomez, Kaiser, and Polosukhin}]{DBLP:conf/nips/VaswaniSPUJGKP17}
Vaswani, A.; Shazeer, N.; Parmar, N.; Uszkoreit, J.; Jones, L.; Gomez, A.~N.;
  Kaiser, L.; and Polosukhin, I. 2017.
\newblock Attention is All you Need.
\newblock In \emph{{NIPS}}.

\bibitem[{Wang et~al.(2022)Wang, Bao, Dong, Bjorck, Peng, Liu, Aggarwal,
  Mohammed, Singhal, Som, and Wei}]{DBLP:journals/corr/abs-2208-10442}
Wang, W.; Bao, H.; Dong, L.; Bjorck, J.; Peng, Z.; Liu, Q.; Aggarwal, K.;
  Mohammed, O.~K.; Singhal, S.; Som, S.; and Wei, F. 2022.
\newblock Image as a Foreign Language: BEiT Pretraining for All Vision and
  Vision-Language Tasks.

\bibitem[{You et~al.(2023)You, Sun, Wang, Chen, Wang, Ayyubi, Chang, and
  Chang}]{DBLP:journals/corr/abs-2305-14985}
You, H.; Sun, R.; Wang, Z.; Chen, L.; Wang, G.; Ayyubi, H.~A.; Chang, K.; and
  Chang, S. 2023.
\newblock IdealGPT: Iteratively Decomposing Vision and Language Reasoning via
  Large Language Models.
\newblock \emph{arXiv}.

\bibitem[{Zhu et~al.(2023{\natexlab{a}})Zhu, Chen, Haydarov, Shen, Zhang, and
  Elhoseiny}]{DBLP:journals/corr/abs-2303-06594}
Zhu, D.; Chen, J.; Haydarov, K.; Shen, X.; Zhang, W.; and Elhoseiny, M.
  2023{\natexlab{a}}.
\newblock ChatGPT Asks, {BLIP-2} Answers: Automatic Questioning Towards
  Enriched Visual Descriptions.
\newblock \emph{arXiv}.

\bibitem[{Zhu et~al.(2023{\natexlab{b}})Zhu, Chen, Shen, Li, and
  Elhoseiny}]{DBLP:journals/corr/abs-2304-10592}
Zhu, D.; Chen, J.; Shen, X.; Li, X.; and Elhoseiny, M. 2023{\natexlab{b}}.
\newblock MiniGPT-4: Enhancing Vision-Language Understanding with Advanced
  Large Language Models.
\newblock \emph{arXiv}.

\end{thebibliography}

\clearpage

\section{Supplementary}
~\\

\begin{figure}[H]
    \centering
    \onecolumn\includegraphics[width=0.8\textwidth]{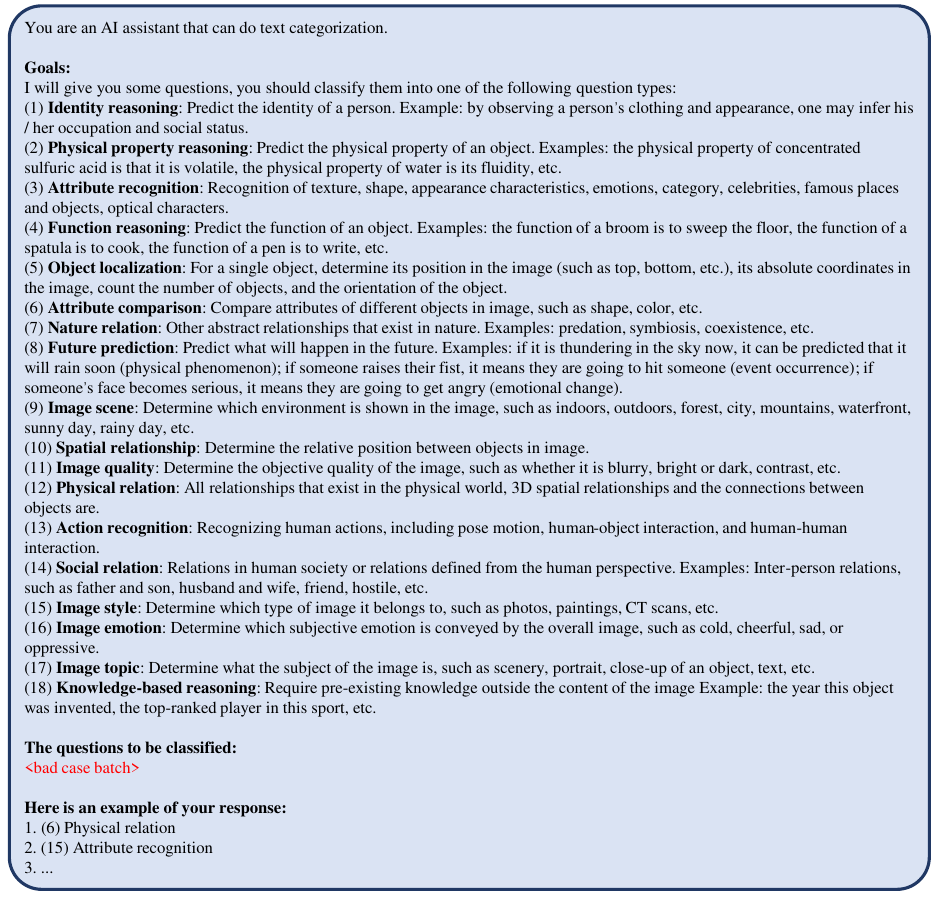}
    \caption{The prompt for question classification. The explanation of each question type is from MMBenchmark \cite{DBLP:journals/corr/abs-2307-06281} (except for Knowledge-based reasoning, which is added by us).}
    \label{fig:Prompt-bad_case_classifiction}
\end{figure}

\begin{figure*}
    \centering
    \includegraphics[width=1\textwidth]{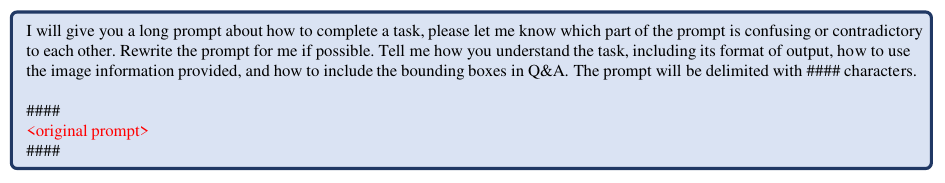}
    \caption{The prompt for modifying confusion and conflict}
    \label{fig:Prompt-confusion_conflict}
\end{figure*}

\begin{figure*}
    \centering
    \includegraphics[width=1\textwidth]{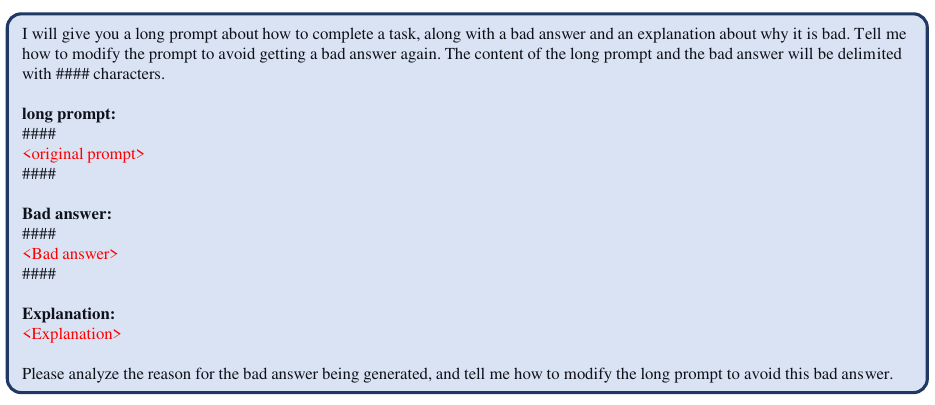}
    \caption{The prompt for avoiding failure case}
    \label{fig:Prompt-bad_case_modify}
\end{figure*}

\begin{figure*}
    \centering
    \includegraphics[width=1\textwidth]{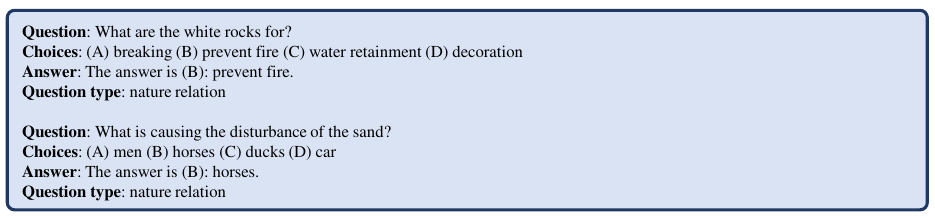}
    \caption{An example of \textless Few shot Example\textgreater for original prompt. }
    \label{fig:Prompt-v0-few-shot}
\end{figure*}

\begin{figure*}
    \centering
    \includegraphics[width=1\textwidth]{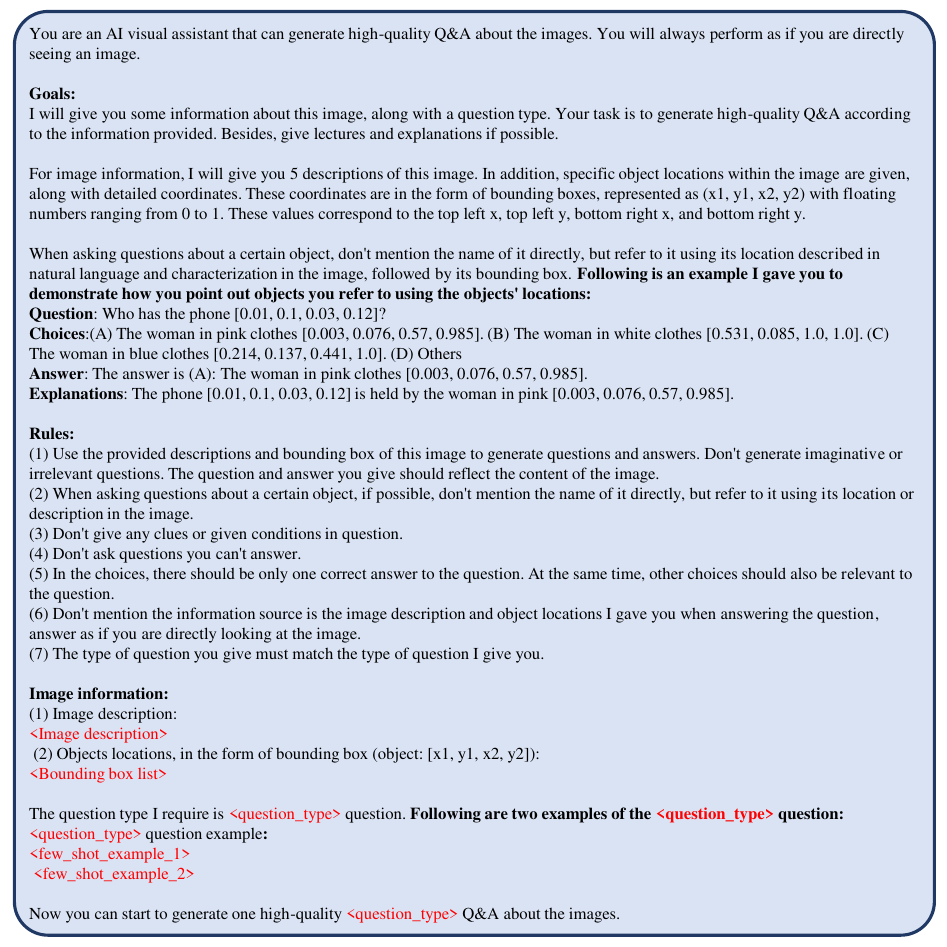}
    \caption{The original version of prompt: The example of \textless Few shot Example\textgreater is as shown in Figure \ref{fig:Prompt-v0-few-shot}. Specifically, we add bbox to reduce the illusion of GPT-4, which can be deleted from the QA without affecting sentence completeness.}
    \label{fig:Prompt-v1}
\end{figure*}

\begin{figure*}
    \centering
    \includegraphics[width=1\textwidth]{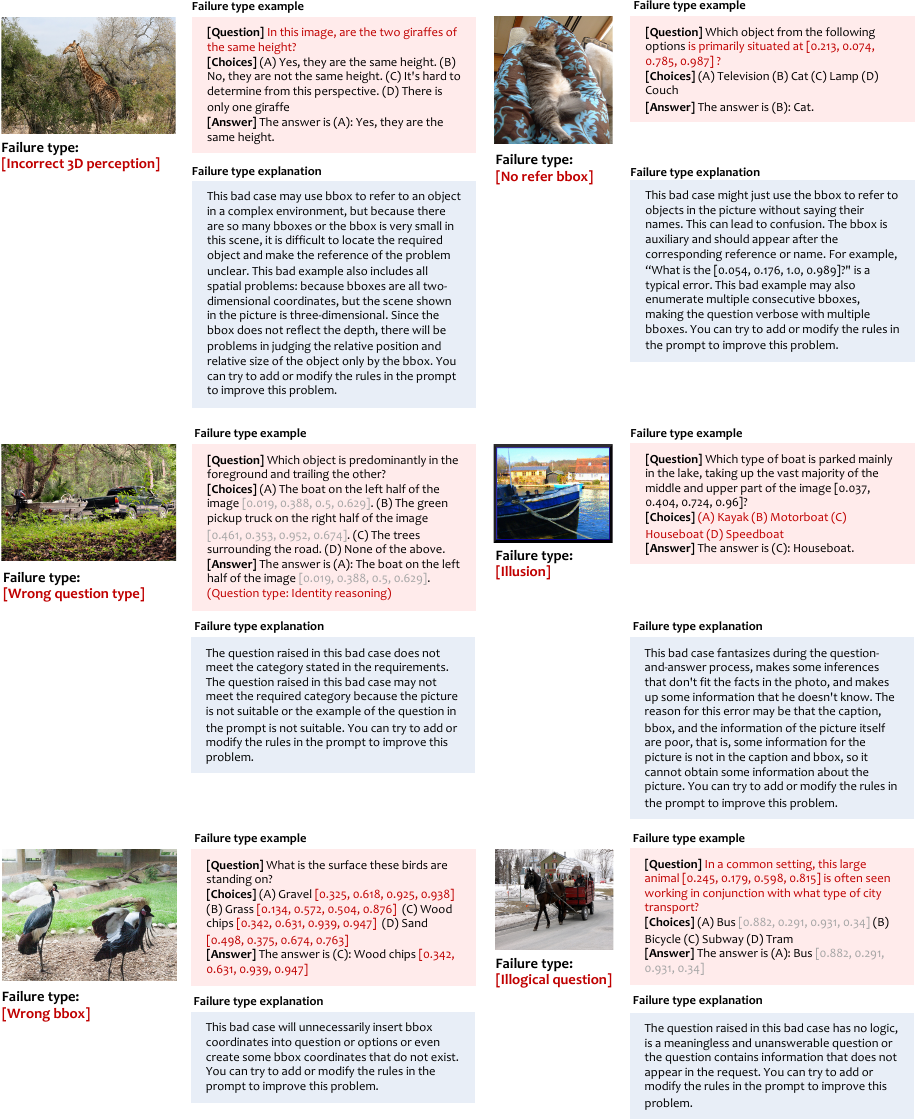}
    \caption{Failure type examples and explanation. The failure type example and its explanation about why it is considered to be failure case are summarized by human from test results.}
    \label{fig:Failure_type_explanation}
\end{figure*}

\begin{figure*}
    \centering
    \includegraphics[width=1.0\textwidth]{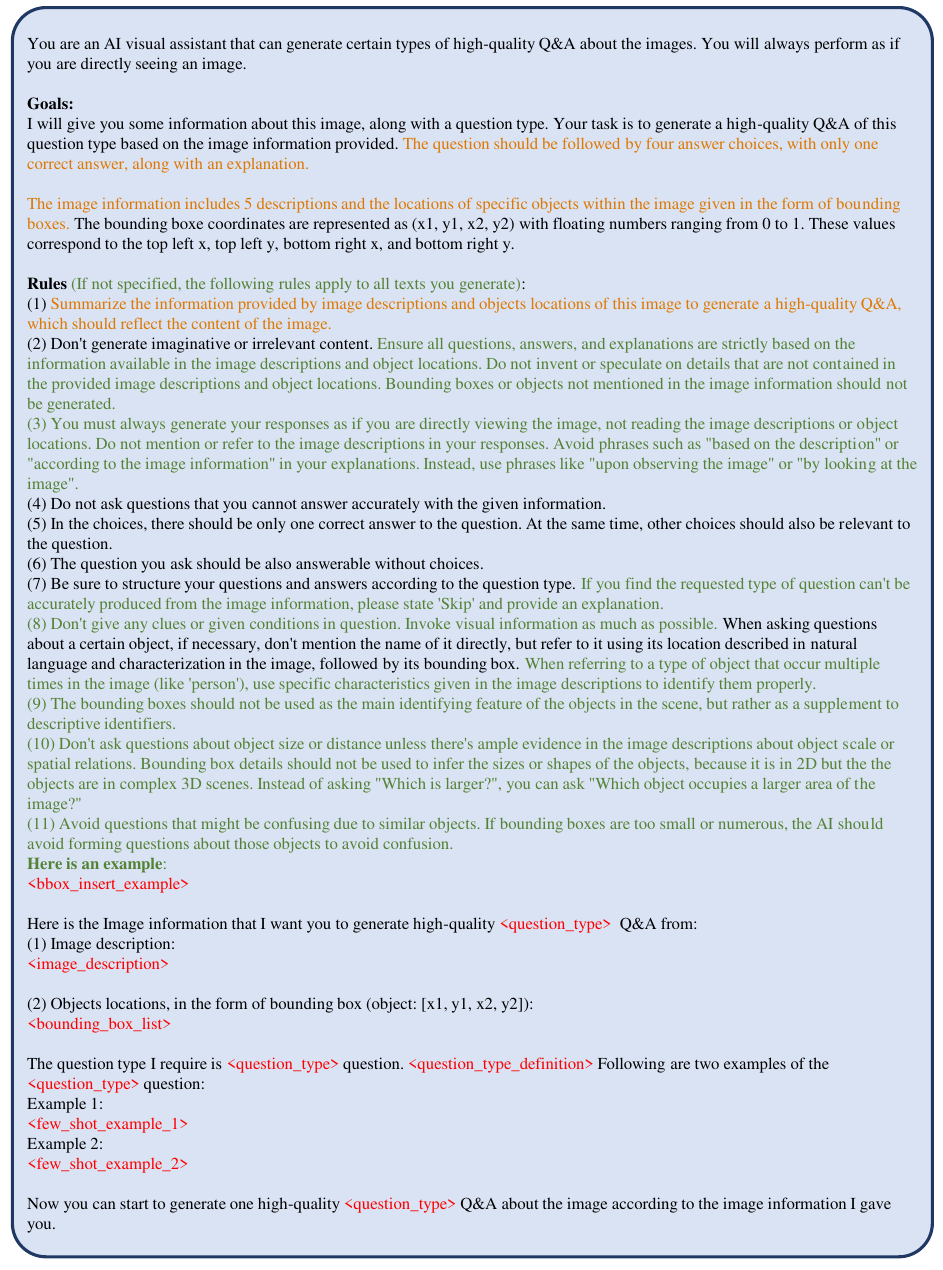}
    \caption{Final version of the prompt. Yellow highlight indicates statement optimization while green highlight represents new additions. The \textless question type definition\textgreater remains consistent with Figure \ref{fig:Prompt-bad_case_classifiction}. The \textless bbox insert example\textgreater is shown in Figure \ref{fig:Prompt-v2-few-shot}.}
    \label{fig:Prompt-v2}
\end{figure*}

\begin{figure*}
    \centering
    \includegraphics[width=1\textwidth]{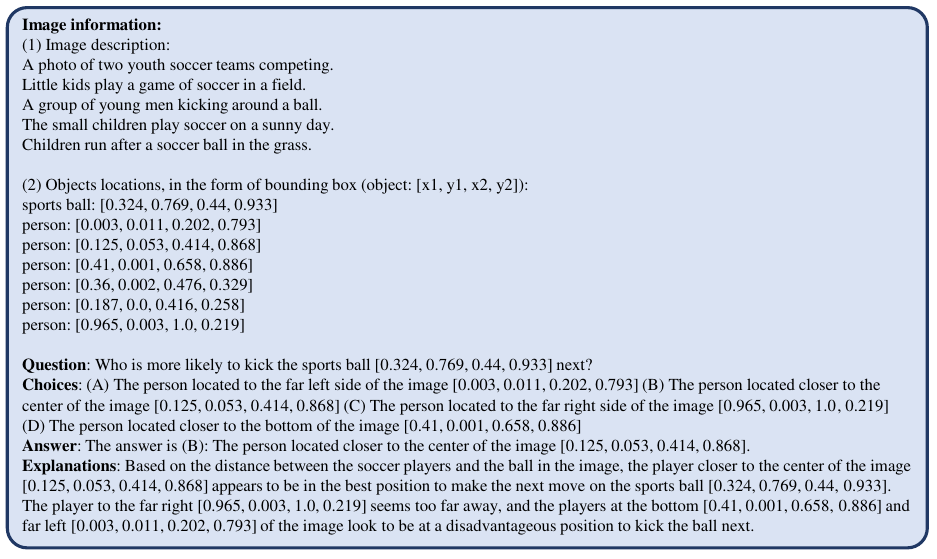}
    \caption{The content of \textless bbox insert example\textgreater in the final prompt}
    \label{fig:Prompt-v2-few-shot}
\end{figure*}

\begin{figure*}
    \centering
    \includegraphics[width=1\textwidth]{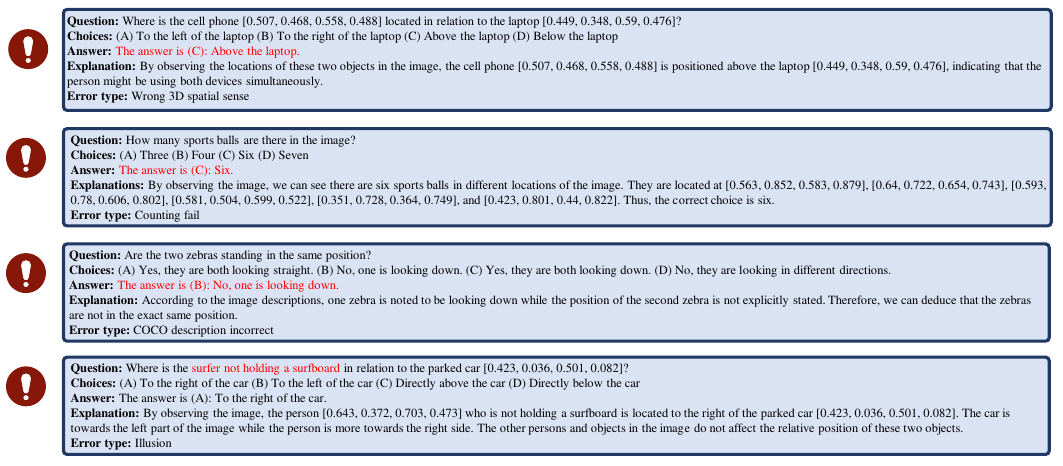}
    \caption{Failure cases remained after optimization}
    \label{fig:Failure_example}
\end{figure*}

\begin{figure*}
    \centering
    \includegraphics[width=1\textwidth]{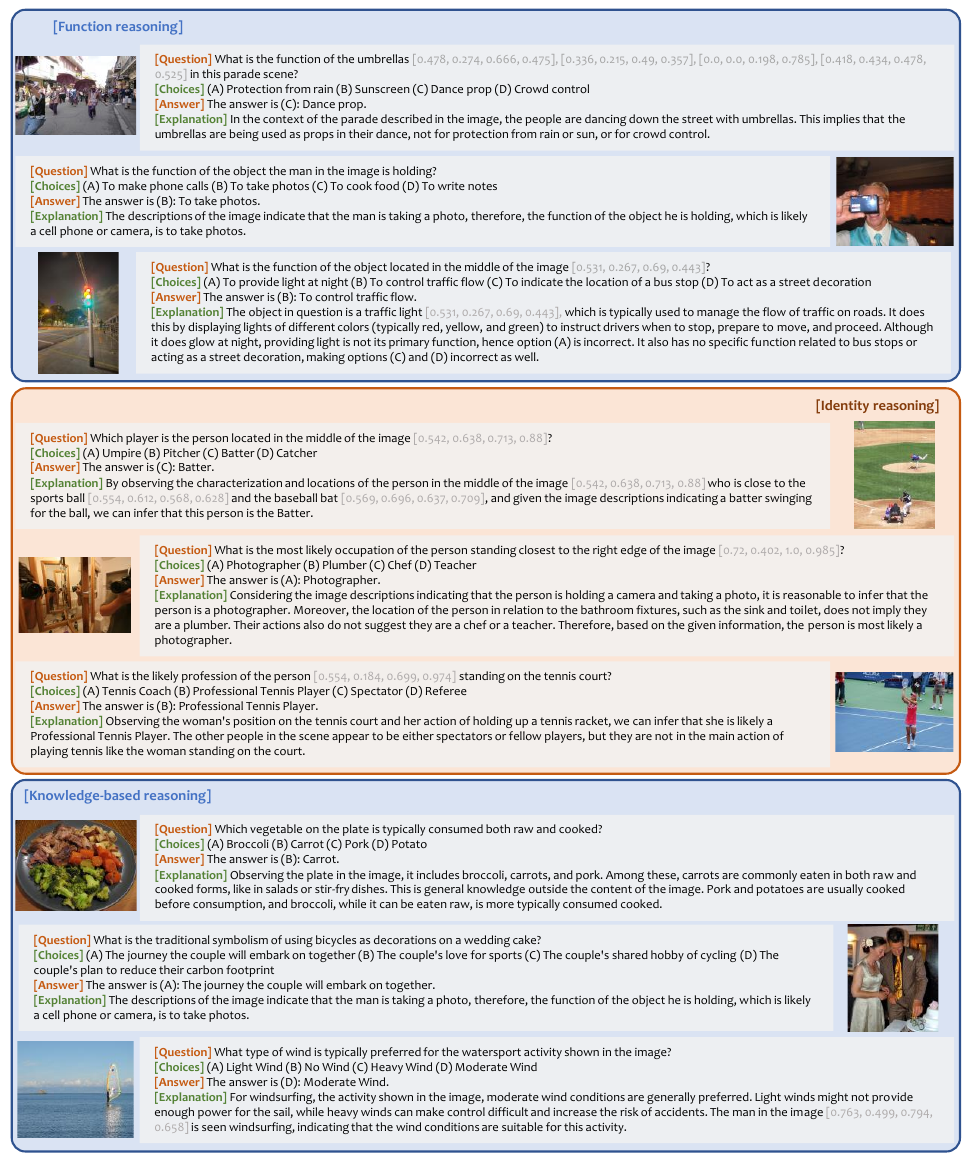}
    \caption{Examples of GPTVQA in function reasoning, identity reasoning, and knowledge-based reasoning}
    \label{fig:Showcase1}
\end{figure*}

\begin{figure*}
    \centering
    \includegraphics[width=1\textwidth]{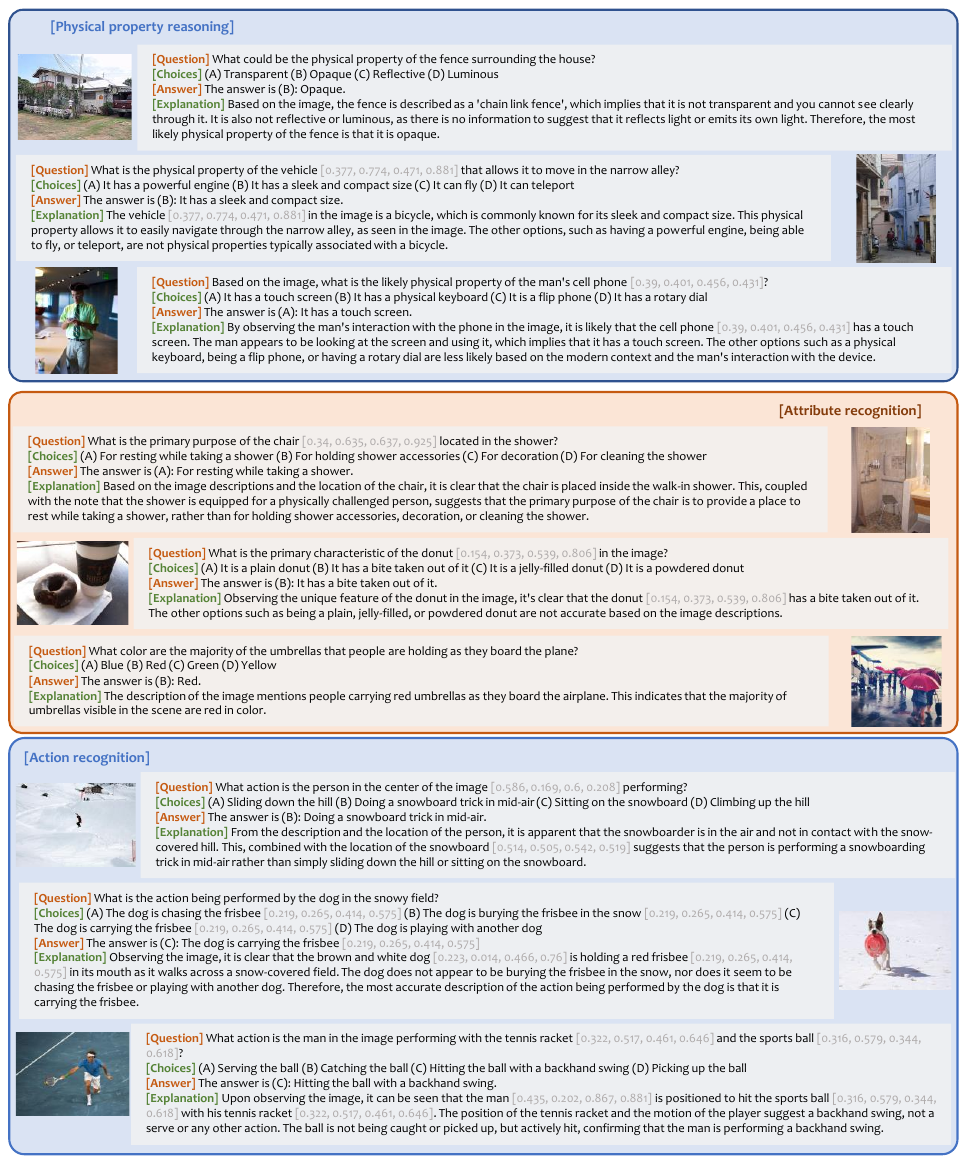}
    \caption{Examples of GPTVQA in physical property reasoning, attribute recognition, and action recognition }
    \label{fig:Showcase2}
\end{figure*}

\begin{figure*}
    \centering
    \includegraphics[width=1\textwidth]{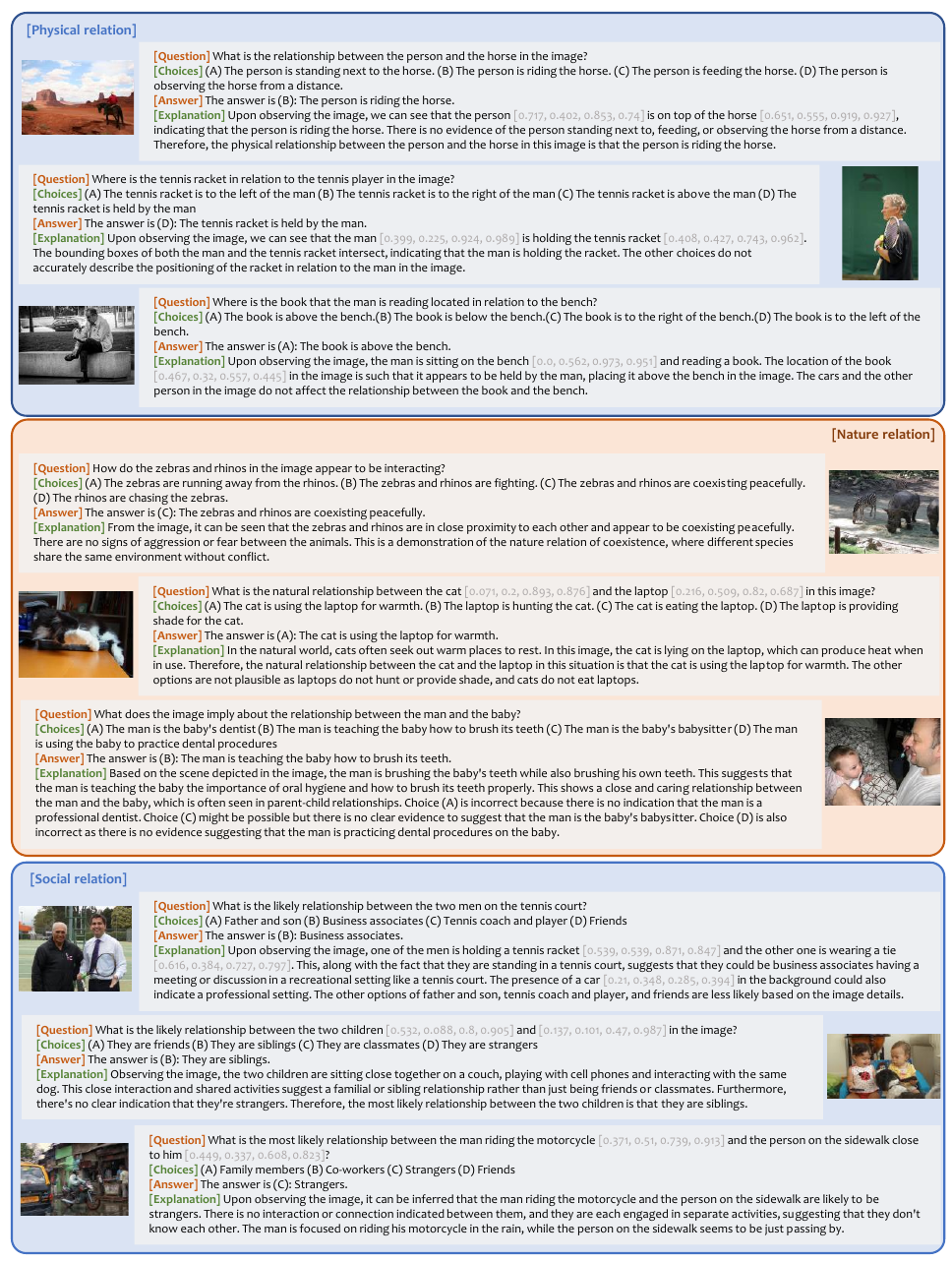}
    \caption{Examples of GPTVQA in physical relation, nature relation, and social relation }
    \label{fig:Showcase3}
\end{figure*}

\begin{figure*}
    \centering
    \includegraphics[width=1\textwidth]{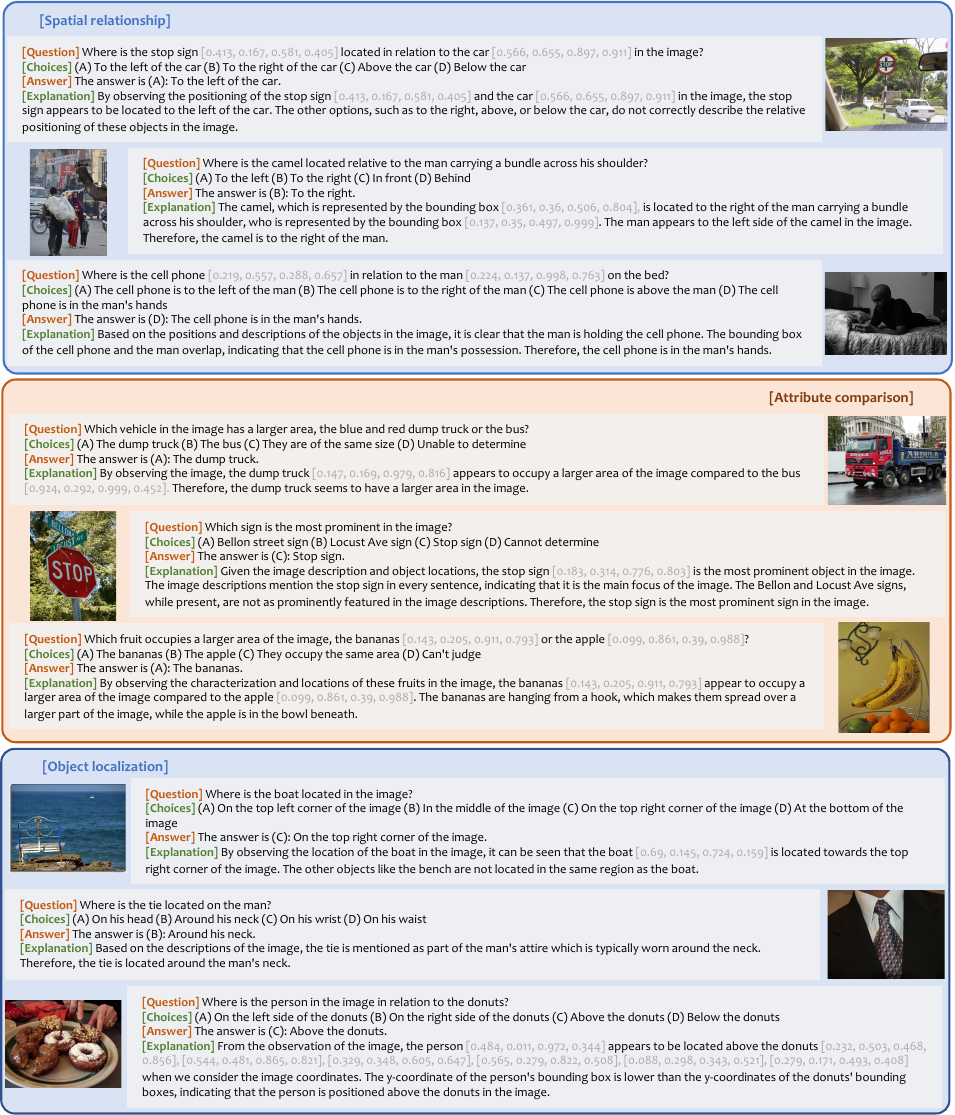}
    \caption{Examples of GPTVQA in spatial relationship, attribute comparison, and object localization}
    \label{fig:Showcase4}
\end{figure*}

\begin{figure*}
    \centering
    \includegraphics[width=1\textwidth]{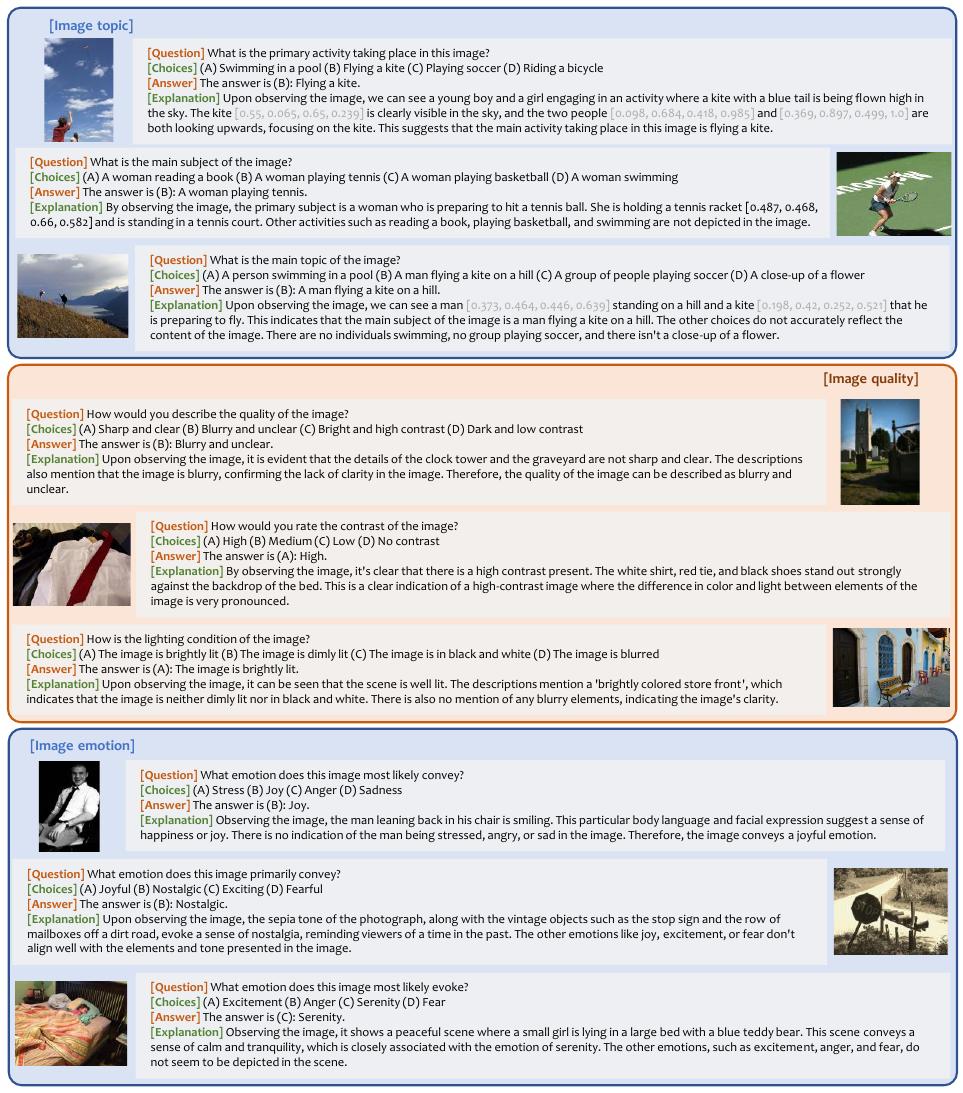}
    \caption{Examples of GPTVQA in image topic, image quality, and image emotion}
    \label{fig:Showcase5}
\end{figure*}

\begin{figure*}
    \centering
    \includegraphics[width=1\textwidth]{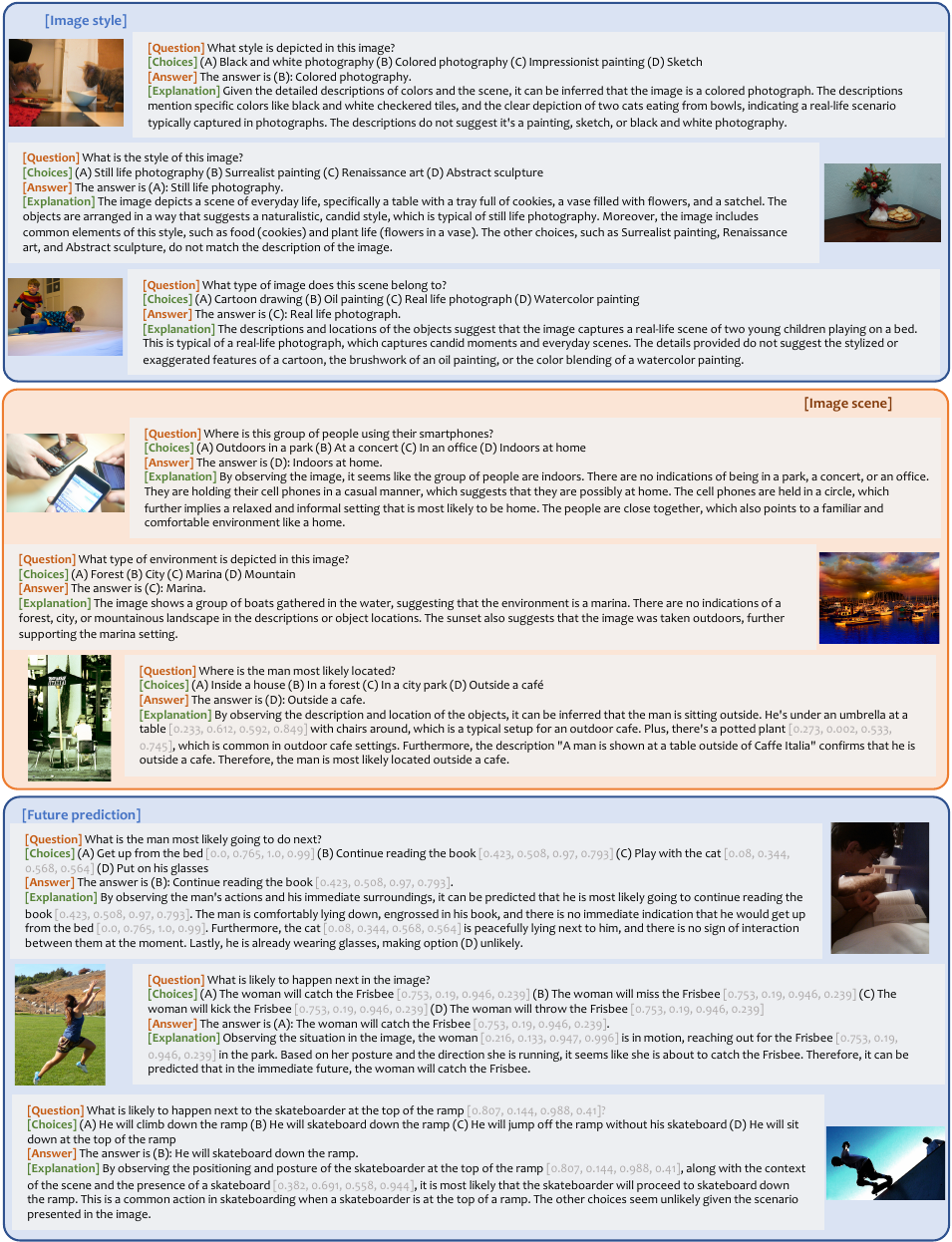}
    \caption{Examples of GPTVQA in image style, image scene, and future prediction}
    \label{fig:Showcase6}
\end{figure*}

\noindent\begin{minipage}{\textwidth}
\noindent\begin{minipage}{0.49\textwidth}
\subsection{Prompts used in MLLM-DataEngine}
The detailed descriptions of the prompts used in each component of the MLLM-DataEngine are as follows: Figure \ref{fig:Prompt-bad_case_classifiction} illustrates the prompt used for categorizing different problems when constructing the bad case pool, which is in last of Model Evaluation phas in the MLLM-DataEngine. In Figure \ref{fig:Prompt-confusion_conflict}, GPT-4 is commanded to modify any confusion or contradictions present in the original prompt and rewrite it accordingly. Figure \ref{fig:Prompt-bad_case_modify} instructs GPT-4 to analyze the root cause of the confusing or contradictory elements and provide recommendations for modifications. Figure \ref{fig:Prompt-v0-few-shot} depicts two selected in-context examples in Adaptive Bad-case Sampling during the query construction phase. Figure \ref{fig:Prompt-v1} presents the initial version of the prompt. Definitions for each failure type in the original prompt are provided in Figure \ref{fig:Failure_type_explanation}. Lastly, Figure \ref{fig:Prompt-v2} denotes the prompt that has been refined by Interactive Prompt Optimization. Fig

\end{minipage}\hfill
\begin{minipage}{0.49\textwidth}

ure \ref{fig:Prompt-v2-few-shot} is the bounding box example in the final prompt. Figure \ref{fig:Failure_example} refers to remaining failure cases in final prompt.

\subsection{Quality Examples in GPTVQA}

High-quality and diverse examples in GPTVQA are demonstrated. Figure \ref{fig:Showcase1} are examples of function reasoning, identity reasoning, and knowledge-based reasoning. Figure \ref{fig:Showcase2} are examples of physical property reasoning, attribute recognition, and action recognition. Figure \ref{fig:Showcase3} are examples of physical relation, nature relation, and social relation. Figure \ref{fig:Showcase4} are examples of spatial relationship, attribute comparison, and object localization. Figure \ref{fig:Showcase5} are examples of image topic, image quality, and image emotion. Figure \ref{fig:Showcase6} are examples of image style, image scene, and future prediction.

\end{minipage}
\end{minipage}


\newpage

\end{document}